\documentclass[12pt]{article}

\usepackage{amsmath}
\usepackage{graphicx,psfrag,epsf,epstopdf}
\usepackage{enumerate}
\usepackage{natbib}
\usepackage{url} 

\usepackage{amssymb}
\usepackage{amsthm}
\usepackage{amsmath}
\usepackage{rotating}
\usepackage{multirow}
\usepackage{booktabs}
\usepackage{xcolor}
\usepackage[hidelinks = true]{hyperref}

\newcommand{\bmx}{{\boldsymbol x}}
\newcommand{\bmX}{{\boldsymbol X}}
\newcommand{\bmy}{{\boldsymbol y}}
\newcommand{\bme}{{\boldsymbol e}}
\newcommand{\bmi}{{\boldsymbol i}}
\newcommand{\bmI}{{\boldsymbol I}}
\newcommand{\bmY}{{\boldsymbol Y}}
\newcommand{\bmZ}{{\boldsymbol Z}}
\newcommand{\bmz}{{\boldsymbol z}}
\newcommand{\bmSS}{{\boldsymbol \Sigma}}
\newcommand{\bmmm}{{\boldsymbol \mu}}
\newcommand{\bmu}{{\boldsymbol u}}
\newcommand{\bmv}{{\boldsymbol v}}
\newcommand{\bmL}{{\boldsymbol L}}
\newcommand{\bmb}{{\boldsymbol b}}
\newcommand{\bmA}{{\boldsymbol A}}
\newcommand{\bmc}{{\boldsymbol c}}
\newcommand\argmin{\operatornamewithlimits{argmin}}

\begin{document}

\title{\LARGE\bf Anomaly detection using data depth: \\ multivariate case}
\author{Pavlo Mozharovskyi \\ {\small LTCI, T{\'e}l{\'e}com Paris, Institut Polytechnique de Paris}\\
Romain Valla \\ {\small LTCI, T{\'e}l{\'e}com Paris, Institut Polytechnique de Paris}}
    
\date{July 8, 2024}
\maketitle

\vspace{5ex}

\begin{abstract}
Anomaly detection is a branch of data analysis and machine learning which aims at identifying observations that exhibit abnormal behaviour. Be it measurement errors, disease development, severe weather, production quality default(s) (items) or failed equipment, financial frauds or crisis events, their on-time identification, isolation and explanation constitute an important task in almost any branch of science and industry. By providing a robust ordering, data depth---statistical function that measures belongingness of any point of the space to a data set---becomes a particularly useful tool for detection of anomalies. Already known for its theoretical properties, data depth has undergone substantial computational developments in the last decade and particularly recent years, which has made it applicable for contemporary-sized problems of data analysis and machine learning.

In this article, data depth is studied as an efficient anomaly detection tool, assigning abnormality labels to observations with lower depth values, in a multivariate setting. Practical questions of necessity and reasonability of invariances and shape of the depth function, its robustness and computational complexity, choice of the threshold are discussed. Illustrations include use-cases that underline advantageous behaviour of data depth in various settings.
\indent\\


\end{abstract}

\newpage

\section{Motivation}

Being applicable in a large variety of domains, anomaly detection increasingly gains popularity among researchers and practitioners. Having been in use since decades, it constitutes a contemporary domain of rapid development to meet growing demand in various areas such as industry, economy, social sciences, \textit{etc.} With large amounts of data recorded in modern applications and constantly present probability of abnormal events, these cannot be identified by operator's hand anymore: automatic procedures are necessary.

It is not the goal of the current article to provide a complete overview of anomaly detection methods, the reader is referred to~\cite{ChandolaBK09}; see also following Sections~\ref{ssec:motiv:difoutl} and~\ref{ssec:motiv:industry} for intuition. Here, a narrower question is in scope: why and how to employ \textit{data depth} for anomaly detection?


\subsection{Difference from outlier detection}\label{ssec:motiv:difoutl}

With two terms ``outlier'' and ``anomaly'' being used by two communities with small overlap, a discussion on their similarity is important.

From statistical point of view, both outlier and anomaly detection focus on identifying atypical observations. Nevertheless, there is a substantial difference in application of methods from these the two groups. First of all, while the term ``outlier detection'' is traditionally used by statisticians, ``anomaly detection'' has been adopted by the machine learning community. As a consequence, (more theoretically oriented) statisticians ``did not need'' and often were unaware of (some of the) anomaly detection methods developed by the machine learning community, while---when searching for practical solutions in applications---machine learners did not find outlier detection methods sufficiently flexible (w.r.t. the data space and shape of the distribution) and scalable (with number of observations and variables). Furthermore, rigorous statistical analysis and inference tools, being often in the center of attention for statisticians, often do not exist for anomaly detection methods, with latter taking frequently form of heuristics.

Indeed, perhaps in the best way the difference between ``outlier'' and ``anomaly'' can be described in application. Given a data set at hand, the task of identification of outliers consists in searching for observations not resembling the majority of the data set. ``Anomaly detection'' approach is more operational and follows rather the philosophy of machine learning. That is, given a training data set, which itself can contain anomalies or not, the task is to construct a rule (training phase) which can assign (on the detection phase) each observation of the space (including the observations of the training set) either to the category of anomalies or normal observations.

This work-flow imposes certain requirements on anomaly detection methodology, \textit{e.g.}, regarding the data set used to learn the anomaly detection rule. Should the rule simply save the entire training data set (this would be the case when directly applying data depth), only part of it, or not at all; should the rule be updated, and how often? Continuing the example with data depth, on the learning phase (again in direct application), training data set should be simply saved in the memory and no computations are to be done. When checking abnormality of a new observation, its data depth shall be computed w.r.t. the (saved) training data set based on which the decision about the observation's abnormality shall be made. To keep the rule scalable (and fitting in limited machine memory), only its subset can be stored instead. In the case of Mahalanobis depth, only parameters (center vector and scatter matrix) need to be saved, and no data at all.

It is important to keep attention on this operational aspect when underlining suitability of data depth for anomaly detection in industrial context in the following Section~\ref{ssec:motiv:industry}, and focus on this aspect later in Section~\ref{sec:advantages}.

\subsection{Industrial context}\label{ssec:motiv:industry}

Regard the following example simulating industrial data. Think of a (potential) production line that manufactures certain items. On several stages of the production process, measurements are taken on each of the items to ensure the quality of the produced pieces. These measurements can be numerous if the line is well automatized, or rare if this is not the case. If---for each item---these measurements can be assembled in a vector (of length $d$), then the item can be represented as a multivariate observation $\bmx$ in an Euclidean space ($\bmx\in\mathbb{R}^d$), and the entire manufacturing process as a data set in the same space ($\{\bmx_1,...,\bmx_n\}\subset\mathbb{R}^d$). 

\begin{figure}[!b]
	\begin{center}
		\includegraphics[trim=3cm .5cm 3cm 1cm,clip=true,width=0.465\textwidth]{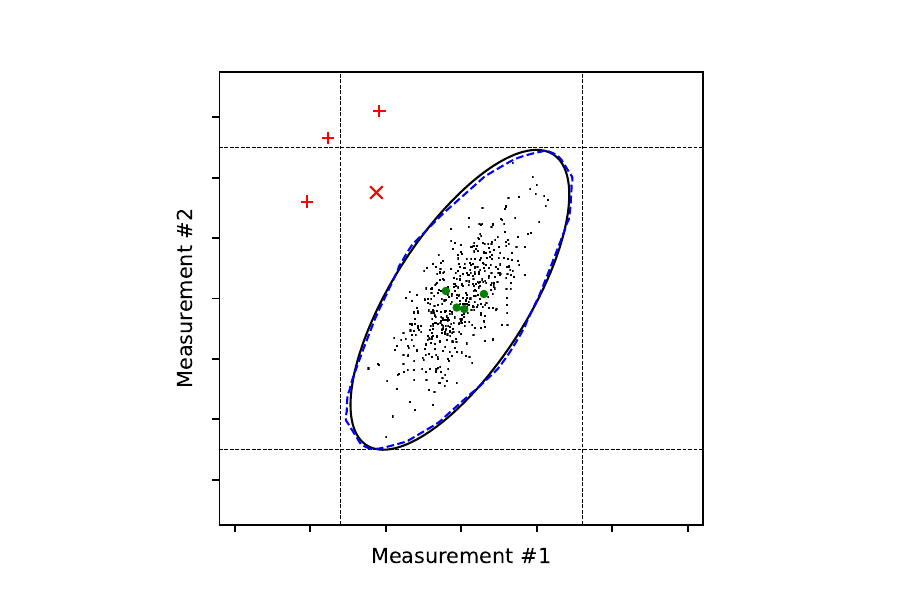}\qquad\includegraphics[trim=3cm .5cm 3cm 1cm,clip=true,width=0.465\textwidth]{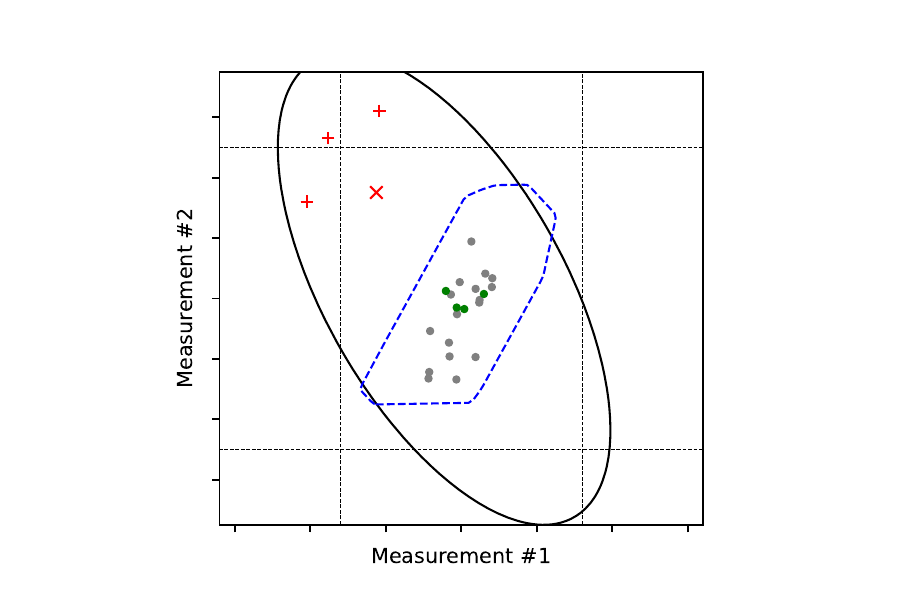}
	\end{center}
	\caption{Four normal observations (green dots) and four anomalies (red pluses and cross); contours of Mahalanobis (black solid) and projection (blue dashed) depths. Left: A sample of $500$ bivariate Gaussian observations (black pixels). Right: $17$ bivariate Gaussian observations (gray dots).}\label{fig:intro}
\end{figure}

Regard Figure~\ref{fig:intro}, left. For visualization purposes, let us restrict to two measurements, whence each produced item is represented by an observation with two variables (=measurements). To construct an anomaly detection rule, a subset of production data is taken as a training set, which can itself contain anomalies or not; this corresponds to the $500$ black pixels, let us denote them $\bmX_{tr}=\{\bmx_{1},...,\bmx_{500}\}\subset\mathbb{R}^2$. $8$ new observations are now to be labeled either as normal observation or as anomalies, namely four green dots (corresponding to normal items), three red pluses and cross (anomalies). While in this bivariate visual case with $d=2$ it is trivially resolved by a simple visual inspection, the task becomes much more complicated once $d$ increases.

The simplest, though still frequently applied approach, is to define validation band for each measurement, \textit{i.e.}, upper and lower bound for each variable: this rule is depicted by black dashed lines parallel to variables' axes and---if well calibrated---allows to identify three out of four anomalies (red pluses) and is computationally extremely fast (computation, as well as following item's production, can even stop after crossing any of the bounds):
\begin{equation}\label{equ:rulebox}
	g_{\text{box}}(\bmx|\bmX_{tr}) = \begin{cases}\text{anomaly\,(=1)}\,,&\,\text{if}\,\bmx\notin\bigcap_{j=1,...,d} (\underline{H}_{j,l_j} \cap \overline{H}_{j,h_j})\,, \\ \text{normal\,(=0)}\,,&\,\text{otherwise}.\end{cases}
\end{equation}
with $l_1,h_1,...,l_d,h_d$ being lower and upper validation bounds (calibrated using $\bmX_{tr}$) for each axis and $\overline{H}_{j,a} = \{\bmy\in\mathbb{R}^d\,|\,\bmy^\top\bme_j\leq a\}$, $\underline{H}_{j,b} = \{\bmy\in\mathbb{R}^d\,|\,\bmy^\top\bme_j\geq b\}$ where $\bme_j$ is the orthant of the $j$th axis. The fourth anomaly (red cross) remains invisible for rule~\eqref{equ:rulebox}.

Obviously, this fourth anomaly can be identified using rule based on Mahalanobis depth $D^{\text{Mah}}$, defined later by~\eqref{equ:depthMah}
\begin{equation}\label{equ:Mahdepth}
	g_{\text{Mah}}(\bmx|\bmX_{tr}) = \begin{cases}\text{anomaly}\,,&\,\text{if}\,D^{\text{Mah}}(\bmx|\bmX_{tr})<t_{\text{Mah},\bmX_{tr}}\,, \\ \text{normal}\,,&\,\text{otherwise}.\end{cases}
\end{equation}
where $t_{\text{Mah},\bmX_{tr}}$ is chosen based on $\bmX_{tr}(=0.075)$ in a way that the Mahalanobis depth contour is largest not to exceed the variable-wise validation bounds. While rule~\eqref{equ:Mahdepth} easily identifies all present anomalies, two aspects shall be taken into account: (i) the training data does not contain anomalies itself and (ii) is large (especially when compared to $d$).

In the beginning of the production process---the phase where diagnostic is particularly important---not many observations are available, but anomalies should still be identified among them; similar situation occurs when produced items are time/resources consuming and are not produced very often. To simulate this situation, regard Figure~\ref{fig:intro}, right. Here, the training data set contains $25$ observations: $19$ being generated from Gaussian distribution (gray dots), $4$ former normal observations (green dots), and the same four anomalies (red pluses and cross). Rule~\eqref{equ:Mahdepth} (with the same threshold $t_{\text{Mah},\bmX_{tr}}$ as before provides misleading ellipse (solid black line) that classifies all anomalies as normal observations. When employing a rule based on projection depth defined later by~\eqref{equ:depthPrj} instead (denoted in blue dashed line), \textit{i.e.},
\begin{equation}\label{equ:projdepth}
	g_{\text{prj}}(\bmx|\bmX_{tr}) = \begin{cases}\text{anomaly}\,,&\,\text{if}\,D^{\text{prj}}(\bmx|\bmX_{tr})<t_{\text{prj},\bmX_{tr}}\,, \\ \text{normal}\,,&\,\text{otherwise}.\end{cases}
\end{equation}
all four anomalies are identified. Furthermore, when all $500$ observations become available (\textit{e.g.}, when the production line work for longer period of time), rule from~\eqref{equ:projdepth} (with the same threshold $t_{\text{prj},\bmX_{tr}}(=0.1575)$) almost coincides with the Mahalanobis depth rule~\eqref{equ:Mahdepth}; see Figure~\ref{fig:intro} (left) again. We refer the reader to Section~\ref{ssec:threshold} for a discussion on the choice of the threshold.

\subsection{Outline of conceptual challenges}

In the rest of the article, after a brief introduction of data depth (Sections~\ref{sec:depths} and~\ref{sec:algorithmic}), challenges connected with its application to anomaly detection are discussed. These can be roughly split in three parts:
\begin{itemize}
	\item Which depth notion to choose in a case at hand (Section~\ref{sec:suitability})?
	\item Why to use data depth, \textit{i.e.}, why and in which cases is it advantageous over existing methods (Section~\ref{sec:advantages})?
	\item How to deal with computational issues when employing data depth (Section~\ref{sec:comptract})?
\end{itemize}
We gather remarks and a disclaimer in Section~\ref{sec:outlook}.

While being destined for practitioners, this article is entirely based on \textit{simulated data}. This is mainly due to four reasons: First, in general enterprises are not willing to share data because of its confidentiality, often for competition reasons. Fortunately, this tendency starts to decrease, which can be witnessed by numerous data challenges, because---depending on the industrial sector---(a) data are getting quickly outdated and much more important is to (quickly) find clues how to treat it or/and (b) enterprise does not have enough internal expertise and searches for external ideas, releasing at least part of their data. Second, industrial cases are normally a result of continuous work (or collaboration) being augmented and labeled over time, often based on several data sets and using \textit{apriori} knowledge of domain experts---a complex situation not necessarily presentable as a simple example. Third, the purpose of data illustrations of this article is to pin cases in which employing data depth based methodology is advantageous, and illustrating it to better degree is more gainful with synthetic data. Fourth, for verification and comparison purposes, a feedback in needed. In any case, there is only very little probability that applicant would encounter exactly the same real data situation (repeated) in practice. (Though all the examples presented in the article are based on simulated data, we shall continue calling them \textit{observations} in what follows.)

\section{What is data depth?}\label{sec:depths}

Data depth is a statistical function that, given a data set $\bmX\subset\mathbb{R}^d$, assigns to each element of the space (where it is defined) a value (usually) between 0 and 1, which characterizes how deep this element is in the data set:
\begin{equation}\label{equ:depth}
	D\,:\,\mathbb{R}^d\times\mathbb{R}^{n\times d} \rightarrow [0,1]\,,\,(\bmx,\bmX) \mapsto D(\bmx|\bmX)\,.
\end{equation}
This element can be an observation that belongs to the data set, or any other arbitrary element of the space, \textit{e.g.}, future observation. Being a function of data, data depth inherits statistical properties of the data set, and thus describes it in one or another manner and can serve many purposes:
\begin{itemize}
	\item it provides natural data-induced ordering on the space, a property not easily extendable beyond univariate data and which is widely used for statistical inference such as classification~\citep{Jornsten04,LiCAL12,LangeMM14a} or testing~\citep{DyckerhoffLP15};
	\item its maximizer(s) is (are) a generalization of median (\textit{i.e.}, robust center) to higher dimensions;
	\item depth contours (constituted of space elements possessing the same depth level) describe data with respect to their location, scatter, and shape, a property that gave rise to the notions of bag-plot~\citep{RousseeuwRT99} and curve box-plot~\citep{MirzargarWK14,LafayeDeMicheauxMV22} being generalizations of the univariate box-plot (see Section~\ref{ssec:depths:contours} below for a discussion on depth contours);
	\item observations with very low depth values are natural candidates for anomalies---a property in the main focus of this article.
\end{itemize}


Let us focus again on the multivariate case, \textit{i.e.}, when the depth is defined in the $d$-variate Euclidean space $\mathbb{R}^d$. Since already in $\mathbb{R}^d$ infinite variety of possible functions fit the definition of data depth from above (including trivial ones, \textit{e.g.}, constant function), requirements (also called postulates) have been put on a depth function to be a proper one. There are two most know sets of such postulates. With the first one formulated by~\cite{Liu90} (for simplicial depth) and generalized to further depths by \cite{ZuoS00a}, we here cite a later one by~\cite{Dyckerhoff04} \citep[see also][for earlier version]{Dyckerhoff02}, because it does not include statistical component (namely behavior for symmetric distribution) and is thus more practical. A depth from~\eqref{equ:depth} should satisfy following postulates:
\begin{itemize}
	\item \textit{Affine invariance}: for any $\bmb\in\mathbb{R}^d$ and any non-singular $d\times d$ matrix $\bmA$ it holds:
	\begin{equation}\label{equ:affine}
		D(\bmA\bmx + \bmb|\bmA\bmX + \bmb) = D(\bmx|\bmX)\,,
	\end{equation}
	where (in slight abuse of notation) $\bmA\bmX + \bmb$ is a shortcut for pre-multiplication with matrix $\bmA$ and adding vector $\bmb$ to each element of $\bmX$.

	This relatively strong but beneficial (as we shall see in Section~\ref{sec:suitability}) postulate can be weakened to orthogonal invariance only, \textit{i.e.} with $\bmA$ being orthogonal matrix.
	\item \textit{Vanishing at infinity}:
	\begin{equation*}
		\lim_{\|\bmx\|\rightarrow\infty}D(\bmx|\bmX)=0\,.
	\end{equation*}
	\item \textit{Monotone relative to deepest point}: for any $\bmx^*$ having maximal depth, \textit{i.e.}, such that $D(\bmx^*|\bmX)=\max_{\bmx\in\mathbb{R}^d}D(\bmx|\bmX)$, and for any $\gamma\in[0,1]$ it holds: 
	\begin{equation}\label{equ:nomonoticity}
		D(\bmx|\bmX) \le D(\bmx^* + \gamma (\bmx - \bmx^*)|\bmX)\,.
	\end{equation}
	This property ensures star-similar shape of the upper-level sets of the depth function. If necessary, it can be strengthened to quasi-concavity of the depth, which would yield convex upper-level regions.
	\item \textit{Upper semicontinuity}: the upper-level sets (called also depth regions) defined as $D_\alpha(\bmX):=\{\bmx\in\mathbb{R}^d\,:\,D(\bmx|\bmX)\ge\alpha\}$ are closed, to make the depth function upper-semicontinuous.
\end{itemize}

Even these postulates are of course not sufficient to restrict a function to a reasonable and even practically useful depth. That is why up to hundred definitions have been attempted, with only a dozen being accepted and used by the community. While below we define six depth functions used in the simulation studies of this article, a longer list of depth notions (neither pretending on completeness) can be provided such as \textbf{Convex hull peeling depth}~\citep{Barnett76,Eddy81}, \textbf{Majority depth}~\citep{Singh91}, \textbf{Zonoid depth}~\citep{KoshevoyM97}, \textbf{$\mathbb{L}_p$ depth}~\citep{ZuoS00a}, \textbf{Spatial depth}~\citep{Serfling02}, \textbf{Expected convex hull depth}~\citep{Cascos07}, \textbf{Geometrical depth}~\citep{DyckerhoffM11}, \textbf{Lens depth}~\citep{LiuM11} generalized in \textbf{$\boldsymbol{\beta}$-skeleton depth}~\citep{YangM17}.

As one shall see from this point on, data depth provides a universal generic methodology for anomaly detection since (almost) any depth notion can be plugged in the rule~\eqref{equ:projdepth}, depending on desired properties and existing (computational and data) limitations. This list of depths is of course not complete, if a complete list can be provided in an article in general with many new notions or modifications of existing ones constantly appearing. Below, we define the five depths employed in this article, while letting the reader to consult provided (and other) references for the rest. 

To underline the practical nature of the article, we shall introduce the depths in their empirical context, \textit{i.e.}, as a deterministic function $D(\bmx|\bmX)$ computing representativeness of any point $\bmx\in\mathbb{R}^d$ w.r.t. a data set $\bmX=\{\bmx_1,...,\bmx_n\}$. Indeed, numerically any empirical distribution can be treated with ties not being as exception. Below, we shall also illustrate the techniques how to turn a data depth notion into asymmetric (on the example of projection depth) or affine-invariant~\eqref{equ:affine} (on the example of simplicial volume depth) depth notion.

\textbf{Mahalanobis depth} is defined as a strictly decreasing transform of the (squared) Mahalanobis distance~\citep{Mahalanobis36} to the mean:
\begin{equation}\label{equ:depthMah}
	D^{\text{Mah}}(\bmx|\bmX) = \frac{1}{1 + (\bmx - \bmmm)^\top\bmSS^{-1} (\bmx - \bmmm)}\,.
\end{equation}
Here, $\bmmm$ stands for the mean vector and $\bmSS^{-1}$ for the inverse of the covariance matrix of the distribution generating $\bmX$.

Taking moment estimates for both quantities, \textit{i.e.}, when $\mathbb{R}^d\ni\bmmm=\frac{1}{n}\sum_{i=1}^n\bmx_i$ and (unbiased) $\bmSS=\frac{1}{n-1}\sum_{i=1}^n (\bmx_i - \bmmm)(\bmx_i - \bmmm)^\top$, as it is the case on Figure~\ref{fig:intro}, results in a fast (having time complexity $O(nd^2 + d^3)$ if $n>>d$) and affine invariant but not robust estimator that can be perturbed even by a single anomaly in $\bmX$ with sufficiently high amplitude. A more robust is the minimum covariance determinant estimator~\citep[MCD;][]{RousseeuwL87,LopuhaaR91} which estimates $\bmmm$ and $\bmSS$ as average and empirical covariance of $\alpha\in(0.5, 1]$ portion of $\bmX$ that minimizes the determinant of $\bmSS$, and can be approximated by a fast stochastic algorithm of~\cite{RousseeuwD99}. Generally speaking, any reasonable quantities can be used instead which estimate $\bmmm$ and $\bmSS$ as center and scatter of $\bmX$-generating law.

Such ``robustification'' may conflict with affine invariance though: approximately computed MCD (or another) estimator will be only approximately affine-invariant and more generally, robust estimation of the covariance matrix, especially with growing dimension $d$, is naturally challenging and can be computationally involving; the reader is referred to some recent works~\citep{CaiL11,AvellaMedinaBFL18}, rather to illustrate the complexity of the task. This issue obstructs on the whole construction of affine-invariant robust depths which achieve affine invariance explicitly involving covariance matrix. This is in particular the case with mentioned above spatial depth; with four out of five depths defined below being implicitly affine-invariant, this issue of depth's affine invariance will be again illustrated during definition of simplicial volume depth.

Mahalanobis depth can be referred to as a parametric depth, since it can be described by $d + \frac{d(d+1)}{2}$ parameters (and thus to perform anomaly detection they are sufficient to be computed and stored on the training phase), but its level contours (being quadratic functionals) take shape of concentric ellipsoids and are thus limited in summarizing data. The following four depths are defined based on data geometry and are non-parametric.

\textbf{Halfspace depth}, also called Tukey or location depth~\citep{Tukey75,DonohoG92}, of $\bmx$ w.r.t. $\bmX$ is defined as the smallest fraction of $\bmX$ that can be contained in a closed halfspace together with $\bmx$. Representing halfspace by the vector orthogonal to its boundary hyperplane leads to the following definition:
\begin{equation}\label{equ:depthHfsp}
	D^{\text{hfsp}}(\bmx|\bmX) = \min_{\bmu\in\mathcal{S}^{d-1}}\frac{1}{n}\sum_{i=1}^n\boldsymbol{1}(\bmx_i^\top\bmu \le \bmx^\top\bmu)\,,
\end{equation}
where $\mathcal{S}^{d-1}$ denotes the unit sphere in dimension $d$ (being $(d-1)$-dimensional surface) and $\boldsymbol{1}(A)$ stands for indicator function that equals $1$ if event $A$ is true and $0$ otherwise. Clearly, since this is an estimator of data depth, the average of the indicator functions estimates empirical mass. Being probably one of the most studied in the literature depth notions, halfspace depth is non-parametric, affine-invariant (without involving estimation of the covariance matrix), and robust: anomalies in $\bmX$ (almost) do not distract depth values of the normal data. These properties come with high computational cost though, with the most efficient implemented algorithm~\citep{DyckerhoffM16} to compute halfspace depth in any dimension $d$ having time complexity $O(n^{d-1}\log n)$; see also \texttt{R}-package \texttt{ddalpha}~\citep{PokotyloMD19,ddalpha} and \texttt{Python} library \texttt{data-depth} for implementation. Further, halfspace depth vanishes (and equals zero) immediately beyond the convex hull of the data.

\textbf{Simplicial volume depth}, also called Oja depth, is defined based on outlyingness measure suggested by~\cite{Oja83}:
\begin{equation}\label{equ:depthOja}
    D^{\text{smpv}}(\bmx|\bmX) = \Bigl(1 + \frac{1}{\binom{n}{d}}\sum_{\begin{matrix}i_1<...<i_d \\ i_1,...,i_d\subset\{1,...,n\}\end{matrix}}\text{vol}_d\bigl(\text{conv}(\{\bmx,\bmx_{i_1},...,\bmx_{i_d}\})\bigr)\Bigr)^{-1}\,,
\end{equation}
with $\text{conv}(A)$ standing for the convex hull of the set $A$ (the smallest convex set containing $A$), and $\text{vol}_d(\cdot)$ being the $d$-dimensional volume, that can be calculated as follows:
\begin{equation}
	\text{vol}_d\bigl(\text{conv}(\{\bmv_1,...,\bmv_{d + 1}\})\bigr) = \frac{1}{d!}\bigl|\det\bigl((1,\bmv_1^\top)^\top,...,(1,\bmv_{d+1}^\top)^\top\bigr)\bigr|\,,
\end{equation}
where $\det(\cdot)$ denotes the determinant of a matrix. The average approximates here expectation of the volume of the set containing $d$ points from $\bmX$ and $\bmx$. Simplicial volume depth is not affine-invariant in its traditional version, and additionally is computationally involved: its time complexity is $O(n^{d}d^3)$.

Simplicial volume depth can be transformed into affine invariant by dividing the mentioned above average over the square root of the determinant of the covariance matrix $\bmSS$:
\begin{equation}\label{equ:depthOjaAi}
    D^{\text{smpv(ai)}}(\bmx|\bmX) = \Bigl(1 + \frac{1}{\binom{n}{d}}\sum_{\begin{matrix}i_1<...<i_d \\ i_1,...,i_d\subset\{1,...,n\}\end{matrix}}\frac{\text{vol}_d\bigl(\text{conv}(\bmx,\bmx_{i_1},...,\bmx_{i_d})\bigr)}{\sqrt{\text{det}(\bmSS)}}\Bigr)^{-1}\,.
\end{equation}
While this technique is sufficient for the simplicial volume depth, in general, affine invariance can be achieved by substituting all points by their whitened versions, $\bmL\bmx_i$ with $\bmSS^{-1} = \bmL^\top\bmL$, as it can be the case for spatial~\citep[see also][]{Serfling10} or lens depth not addressed here in detail.

\textbf{Projection depth}~\citep{ZuoS00a} is defined as a strictly decreasing transform of the projected outlyingness~\citep{Stahel81,Donoho82}, which idea can be briefly formulated as follows: a point is outlying if it is outlying in projection on at least one direction. Naturally, by definition, projection depth is thus ``searching'' for this outlyingness-proving direction:
\begin{equation}\label{equ:depthPrj}
	D^{\text{prj}}(\bmx|\bmX) = \Bigl(1 + \max_{\bmu\in\mathcal{S}^{d-1}}\frac{\bigl|\bmx^\top\bmu - \text{med}(\bmX^\top\bmu)\bigr|}{\text{MAD}(\bmX^\top\bmu)}\Bigr)^{-1}\,,
\end{equation}
where (in slight abuse of notation) $\bmX^\top\bmu$ is a shortcut for projecting $\bmX$ on $\bmu$ resulting in $\bmX^\top\bmu=\{\bmx_1^\top\bmu,...,\bmx_n^\top\bmu\}$ and \text{med} and \text{MAD} stand for (univariate) median and median absolute deviation from the median, respectively (for a univariate set $Y$, $\text{MAD}(Y)=\text{med}\bigl(\bigl|Y - \text{med}(Y)\bigr|\bigr)$). Any other robust estimators of univariate location and scale can be used instead, of course.
Projection depth is affine invariant (implicitly, in that resembling the halfspace depth), (highly) robust \citep[asymptotic breakdown point of its median equals $0.5$;][]{Zuo03}, and (different to the halfspace depth) it is positive on the entire $\mathbb{R}^d$ once the depth is well defined. When describing the data, its contours retain certain degree of symmetry. To improve on this, \textbf{asymmetric projection depth}~\citep{Dyckerhoff04} has been proposed:
\begin{equation}\label{equ:depthAPrj}
	D^{\text{prj(as)}}(\bmx|\bmX) = \Bigl(1 + \max_{\bmu\in\mathcal{S}^{d-1}}\frac{\bigl(\bmx^\top\bmu - \text{med}(\bmX^\top\bmu)\bigr)_+}{\text{MAD}_+(\bmX^\top\bmu)}\Bigr)^{-1}\,,
\end{equation}
where $(a)_+=\max\{a,0\}$ and $\text{MAD}_+$ is the median of the positive deviations from the median.

\textbf{Simplicial depth} is defined as the fraction of the---based on $d+1$ data points---simplices that contain $\bmx$~\citep{Liu90}:
\begin{equation}\label{equ:depthSmpl}
    D^{\text{smp}}(\bmx|\bmX) = \frac{1}{\binom{n}{d}}\sum_{\begin{matrix}i_1<...<i_{d+1} \\ i_1,...,i_{d+1}\subset\{1,...,n\}\end{matrix}}\boldsymbol{1}\bigl(\bmx\in\text{conv}(\{\bmx_{i_1},...,\bmx_{i_{d+1}}\})\bigr)\,.
\end{equation}
It is (implicitly) affine-invariant and robust, but approximating the average belongingness to a simplex involves high computational time complexity $O(n^{d+1}d^3)$.
Different to the five depth notions, previously mentioned in this section, simplicial depth has non-convex contours; the same as the halfspace depth, it vanishes immediately beyond the convex hull of the data set $\bmX$.

\section{Algorithmic aspects}\label{sec:algorithmic}

This section provides information about computational aspects of data depths, which are important for anomaly detection. Section~\ref{ssec:depths:comp} proposes a taxonomy of data depth based on their calculation properties. Section~\ref{ssec:depths:contours} refers to use of depth-trimmed regions for anomaly detection.

\subsection{Computational taxonomy}\label{ssec:depths:comp}

Several taxonomies have been proposed to categorize existing data depth notions into groups, mostly by construction mechanism of depths; let us cite two of them. \cite{ZuoS00a} divide multivariate data depth notions in four types: \textit{A} (depths based on closeness to the sample), \textit{B} (depths based on distance to the sample), \textit{C} (depth based on outlyingness measure), and \textit{D} (depths based on ``tailedness'' measured using a class of closed subsets). \cite{Mosler13} categorizes depths in those based on distances (including $\mathbb{L}_p$, Mahalanobis, simplicial volume, and projection depths), weighted mean depths, and depths based on halfspaces and simplices.

\cite{Dyckerhoff04} introduced a class of depths that satisfy the (weak) projection property: they can be computed as minimum of depths in univariate projections. Including zonoid, expected convex hull, geometrical, Mahalanobis, halfspace, projection depths, this class is very important since it allows to develop efficient algorithms based on computation of univariate depths only. Further, \cite{DyckerhoffM11} \citep[see also][]{DyckerhoffM12} define a subclass of these depths---depths defined via (convex) weighted-mean trimmed regions, that include zonoid, expected convex hull, and geometrical depths.

In the current article, we suggest a novel approach---the \textit{computational taxonomy}. Having undergone substantial theoretical developments during last $30$ years data depth has options to offer, and in many situations a depth notion (or its modification) that satisfies expected theoretical properties (important for application of interest and thus potential data generating distribution) can be found or constructed by modification of an existing notion. Today, in the era of big data, computational properties gain even more importance, and the choice of depth can be restricted by computational time and/or resources.

In what follows, we address exact computation (\textit{i.e.} computing depth of observation $\bmx$ with respect to data set $\bmX$) of all above mentioned depths in view of three axes: computational time complexity, affine invariane, and robustness; see Table~\ref{tab:complexity}. While only for several depths the complexity is either proved or obvious, the below summary is relying on complexities of developed algorithms (and their implementations) based on the preceding decades of research.

\begin{table}
	\caption{\label{tab:complexity}Computational taxonomy of data depth along three axes: computational time complexity, affine invariance, and robustness; depth notions in \textit{italics} are robust.}
	
	\centering
	
	\indent\\
	
	\begin{tabular}{cc|c|c}
		& & Exponential & Polynomial \\
		& & time & time \\ \hline
		\multirow{7}{*}{\begin{sideways}Affine-\,\,\,\end{sideways}} & \multirow{7}{*}{\begin{sideways}invariant\,\,\,\end{sideways}} & \textit{convex hull peeling} & zonoid \\ 
		& & \textit{majority} & Mahalanobis \\
		& & expected convex hull &  \\
		& & geometrical &  \\
		& & \textit{halfspace} &  \\
		& & \textit{projection} &  \\
		& & \textit{simplicial} &  \\ \hline
		\multirow{6}{*}{\begin{sideways}\,Not affine-\,\end{sideways}} & \multirow{6}{*}{\begin{sideways}\,\,invariant\,\,\end{sideways}} & simplicial volume & $\mathbb{L}_2$ \\ 
		& & & \textit{spatial} \\
		& & & \textit{lens} \\
		& & & \\
	\end{tabular}
\end{table}

Several remarks are in order here:
\begin{itemize}
	\item Exponential complexity for halfspace depth has been shown by~\cite{JohnsonP78}, this of convex hull peeling depth is mentioned in its computing algorithm~\citep{BarberDH96}, while projection depth is believed to be so as well~\citep{LiuZ14}. Complexities of majority, $\mathbb{L}_p$, spatial, lens, Mahalanobis, simplicial volume and simplicial depths follow from definitions being combinatorial sums of certain order.
	\item Zonoid depth can be formulated as a linear programming task~\citep{DyckerhoffKM96}, which can be solved in polynomial time using the interior point method~\citep{Karmakar84}.
	\item Properly speaking, robustness properties of the $\mathbb{L}_p$-depth depend on the chosen norm degree $p$; we mention only better studied $\mathbb{L}_2$-depth here.
	\item No exact algorithms have been designed for computing expected convex hull and geometrical depths, which can be trivially done when optimizing these depths by searching through contours (exploiting the mentioned above monotonicity property~\eqref{equ:nomonoticity}). Algorithm for computing such contours possesses exponential time complexity~\citep{BazovkinM12}.
	\item The non-existent entry in Table~\ref{tab:complexity} is a depth being simultaneously affine-invariant, robust, and having polynomial time complexity. Making spatial or lens depth affine invariant is possible by using covariance matrix, as it is the case above for affine-invariant simplicial volume depth $D^{\text{smpv(ai)}}$~\eqref{equ:depthOjaAi}. On the other hand, no polynomial-time algorithm exists for computing such a matrix exactly in a robust way without further assumptions.
\end{itemize}

\subsection{A word about depth contours}\label{ssec:depths:contours}

Being a function defined on $\mathbb{R}^d$, data depth gives rise to depth-trimmed central regions, mentioned already above (when listing depth postulates) and defined---for the data set $\bmX$ and depth level $\alpha\in[0,1]$---as:
\begin{equation*}
	D_\alpha(\bmX):=\{\bmx\in\mathbb{R}^d\,:\,D(\bmx|\bmX)\ge\alpha\}\,.
\end{equation*}
These regions describe data with respect to their location, scatter, and shape and can be used for insightful visualization in dimensions $2$ and $3$; see, \textit{e.g.}, \cite{LiuPS99,BazovkinM12,LiuMM19} to name but a few. Further, they can be used to define anomaly detection rule, for a properly chosen depth level $\alpha(t_{\bmX_{tr}})$:
\begin{equation}\label{equ:regionrule}
	g_{\text{reg}}(\bmx|\bmX_{tr}) = \begin{cases}\text{anomaly}\,,&\,\text{if}\,\,\bmx\notin D_{\alpha(t_{\bmX_{tr}})}(\bmX_{tr})\,, \\ \text{normal}\,,&\,\text{otherwise}.\end{cases}
\end{equation}

Seeming easy to write and implement, rule~\eqref{equ:regionrule} conceals substantial computational difficulties. These lie in calculating the depth contours themselves and can be expressed as follows. Mahalanobis depth contours are simple ellipsoids and do not reflect properly the shape of the data. Algorithms computing contours of depth notions attractive from anomaly detection point of view have time complexity growing exponentially with space dimension, and for a number of depths (\textit{e.g.}, lens or simplicial depth) are unknown to the literature. For certain depths (\textit{e.g.}, halfspace or zonoid depth) only necessary part of the contour---\textit{e.g.}, where (majority of) anomalies are expected---can be computed, but such an approach is difficult to justify in practice, while it still becomes intractable with growing space dimension. Indeed, taking into account the currently developed algorithmic basis, it is much more practical to first compute depth of an observation, and then check whether it belongs to depth region (of normal data) by comparing with a threshold.

A possible solution to the above mentioned issue could be development of approximating algorithms for depth contours relying on a small number of (possibly simple) surfaces elements.

\section{Suitability for anomaly detection}\label{sec:suitability}

With the main task of this section being illustration of the very mechanism of application of data depth to the anomaly detection task, we focus on the task of identifying anomalies in the training data. This allows to keep the exposition simple preserving the main difficulty of unsupervised anomaly detection, \textit{i.e.}, to construct a rule not perturbed by the anomalies in the training data. It is from Section~\ref{sec:advantages} that we follow the machine-learning framework and consider two separate data sets for training and testing.

To perform anomaly detection in practice, two natural questions arise first: (i) which depth to choose how and (ii) how (exactly) to proceed?


\subsection{Choice of the depth function}\label{ssec:depthchoice}
\cite{MoslerM22} discuss that when applying data depth, choice of a suitable depth notion is crucial, and usually consists in finding a compromise between its statistical and computational properties. To provide insights on this, let us take a look at the following (somewhat general) simulated example.

Given a training data set $\bmX_{tr}$ of $100$ observations, where $90$ of them are generated from bivariate normal distribution $\mathcal{N}\Bigl((1, 1)^\top,\,\bigl(
 \begin{smallmatrix}
  1 & 1\\
  1 & 2
 \end{smallmatrix}
\bigr)\Bigr)$
with the remaining $10$ stemming also from bivariate normal distribution but with $36$ times smaller covariance and located at the previous mean shifted $2.5$ in direction of the second principal component of normal data:
$\mathcal{N}\Bigl((3.181, -0.222)^\top,\,\bigl(
 \begin{smallmatrix}
  1 & 1\\
  1 & 2
 \end{smallmatrix}
\bigr) / 36\Bigr)$. This example constitutes a rather favorable case, \textit{e.g.}, the $10$ anomalies can be identified by visual inspection (which is of course impossible in higher dimensions).

\begin{figure}[!t]
	\begin{center}
	\begin{tabular}{cc}
		Projection depth & Halfspace depth \\
		\includegraphics[trim=22.5cm 0cm 27.5cm 7.5cm,clip=true,scale=.09]{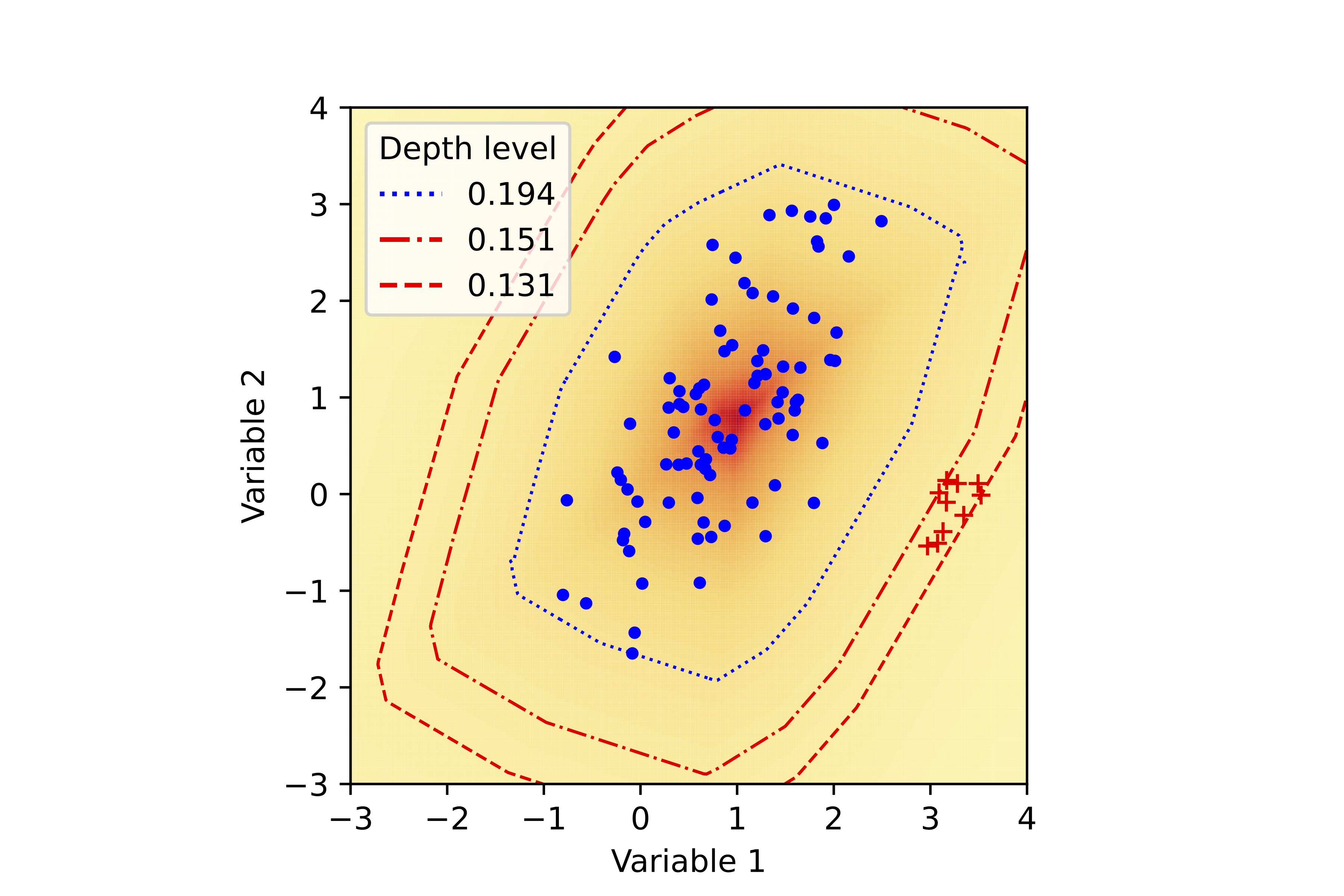} & \includegraphics[trim=17.5cm 0cm 15cm 7.5cm,clip=true,scale=.09]{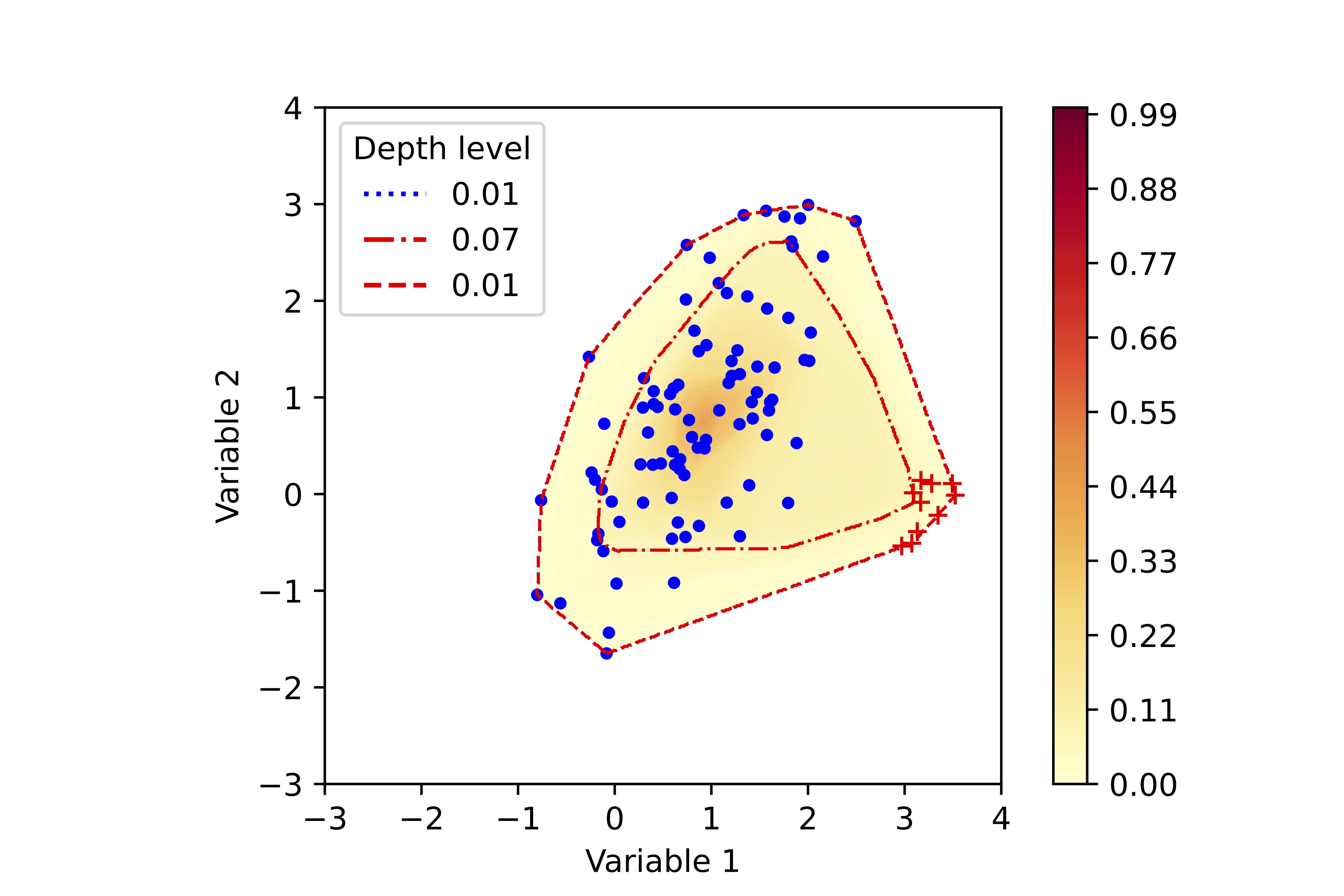}
	\end{tabular}
	\end{center}
	\caption{$90$ observations stemming from bivariate normal distribution (blue dots) contaminated with $10$ observations (red pluses). Depth contours at three levels: minimal depth of normal observations (blue dotted line), maximal depth of $10$ anomalies contaminating the training sample (red dash-dotted line), minimal depth of $10$ anomalies contaminating the training sample (red dashed line). Left: depth values (in color) for projection depth. Right: depth values (in color) for the halfspace depth (with white corresponding to zero). Color scale for both plots is depicted on the right side.}\label{fig:ability1:prjhfsp}
\end{figure}

First, consider projection depth~\eqref{equ:depthPrj} for anomaly detection. For the training sample mentioned right above, Figure~\ref{fig:ability1:prjhfsp} (left) plots depth values $D^\text{\text{prj}}(\bmx|\bmX_{tr})$ of the entire Euclidean space (\textit{i.e.}, for all $\bmx\in\mathbb{R}^2$, see the depth scale on right side of the figure), as well as three depth contours for minimal depth of normal observations, and maximal and minimal depth of anomalies. As it is seen visually, based on the provided training set of $100$ observations (including $10$ anomalies), rule~\eqref{equ:projdepth} allows to detect anomalies when choosing a reasonable threshold $t_{\text{prj},\bmX_{tr}}$ (between $0.151$ and $0.194$ here).

\begin{figure}[!b]
	\begin{center}
		\includegraphics[trim=0.35cm 0cm 1.5cm 1cm,clip=true,width=0.49\textwidth]{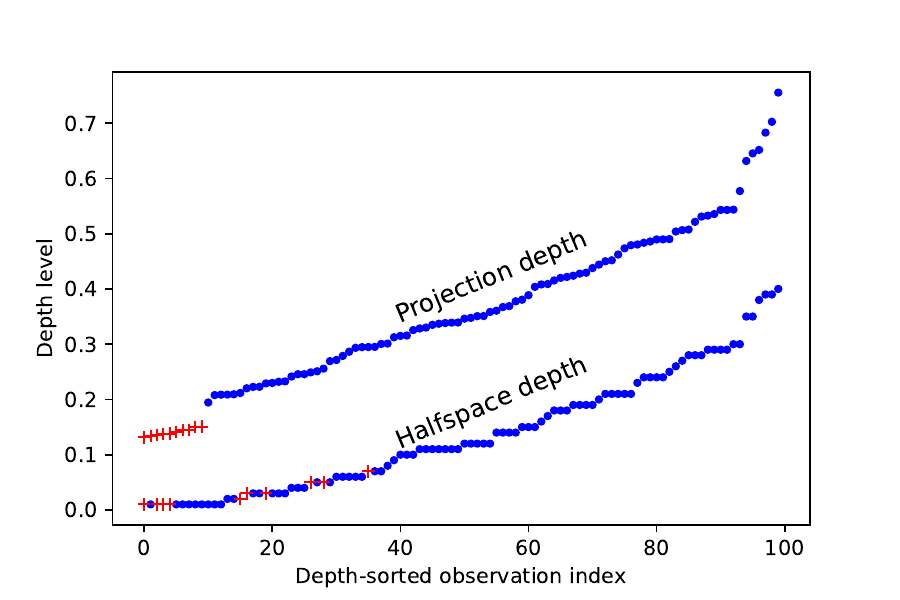}\,\includegraphics[trim=0.35cm 0cm 1.5cm 1cm,clip=true,width=0.49\textwidth]{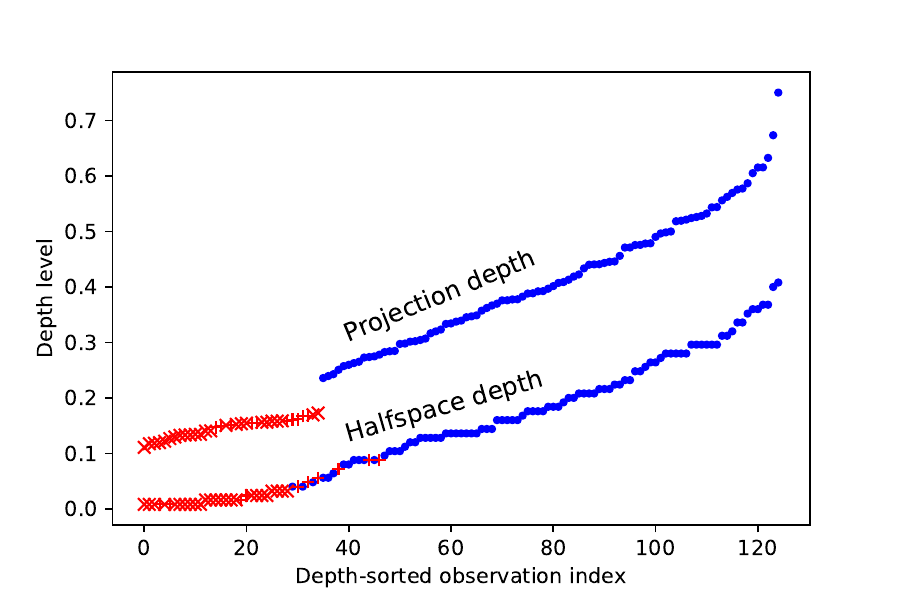}
	\end{center}
	\caption{Ordered depth values for projection and halfspace depth, with normal data corresponding to blue points and anomalies depicted with red pluses and crosses. Left: For $100$ observations of the training data from first example in Section~\ref{sec:suitability}. Right: For $125$ observations of the training data from second example in Section~\ref{sec:suitability}.}\label{fig:ability1:depths}
\end{figure}

Second, consider employing halfspace depth~\eqref{equ:depthHfsp} in the same way. Plot of Figure~\ref{fig:ability1:prjhfsp} (right), which depicts both depth values and contours according to the same logic, shows that correct detection of anomalies is rather impossible, which is in part due to immediate vanishing of the halfspace depth beyond the convex hull of the data. 
A number of works attempted to improve on this last issue, \textit{e.g.}, \cite{EinmahlLL15,RamsayDL19,NagyD21} to name a few, with halfspace mass~\citep{ChenTWH15} being a computationally tractable alternative lacking affine invariance property though.

\begin{figure}[!t]
	\begin{center}
	\begin{tabular}{cc}
		\quad Simplicial volume depth & Simplicial depth \\
		\includegraphics[trim=11cm 0cm 13.75cm 3.75cm,clip=true,scale=.177]{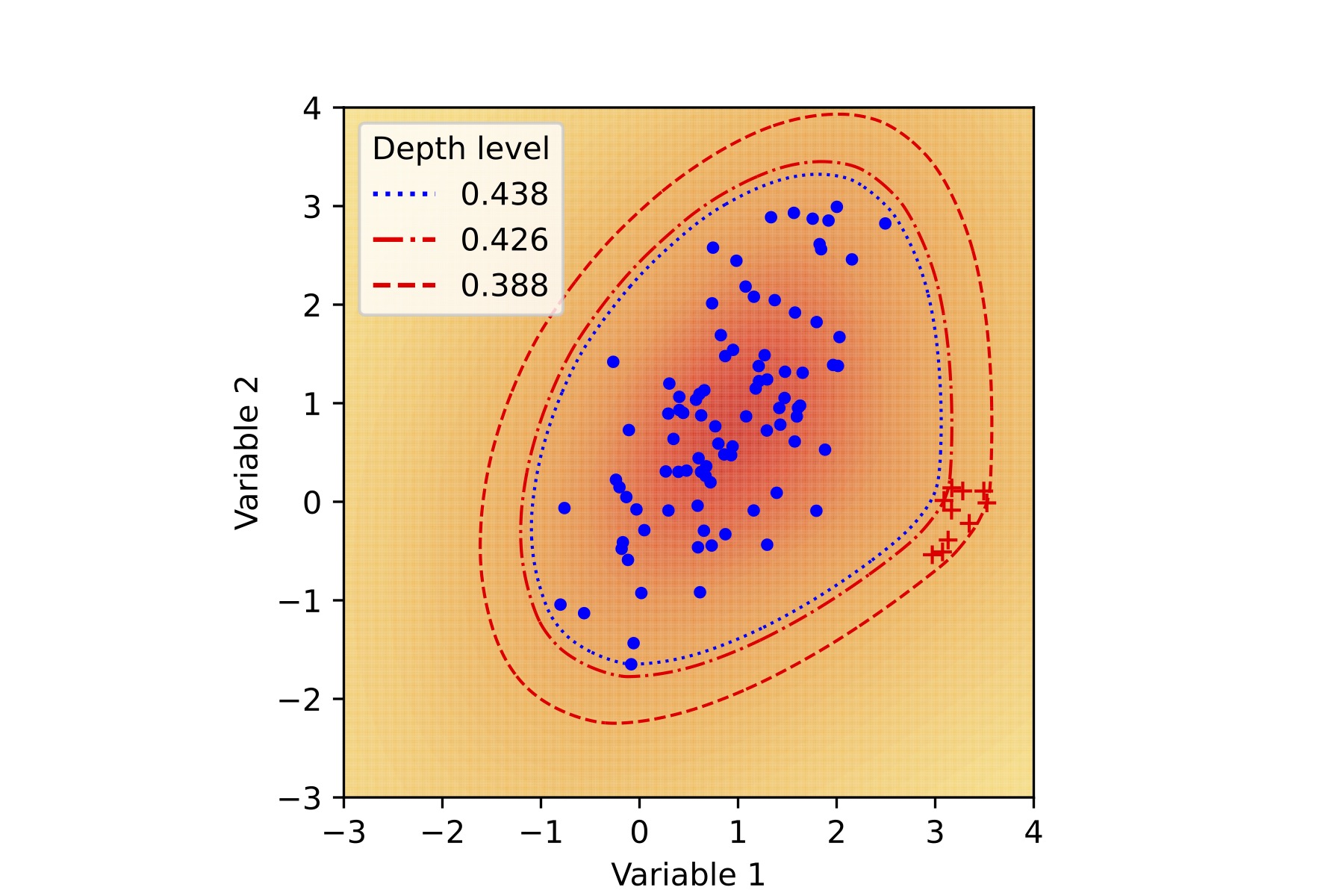} & \includegraphics[trim=17.5cm 1.25cm 15cm 7.5cm,clip=true,scale=.09]{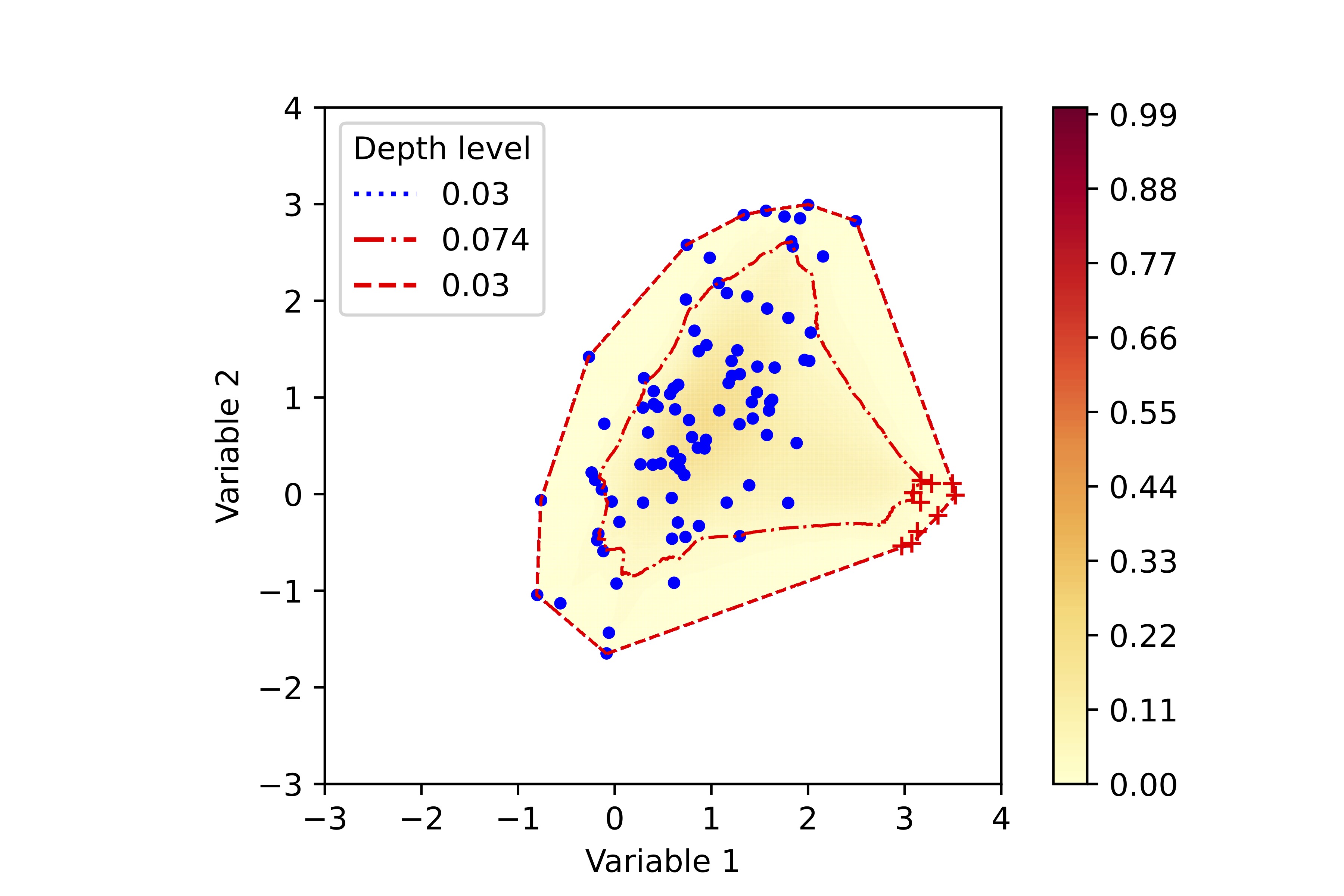}
	\end{tabular}
	\end{center}
	\caption{$90$ observations stemming from bivariate normal distribution (blue dots) contaminated with $10$ observations (red pluses). Depth contours at three levels: minimal depth of normal observations (blue dotted line), maximal depth of $10$ anomalies contaminating the training sample (red dash-dotted line), minimal depth of $10$ anomalies contaminating the training sample (red dashed line). Left: depth values (in color) for simplicial volume depth. Right: depth values (in color) for simplicial depth (with white corresponding to zero). Color scale for both plots is depicted on the right side.}\label{fig:ability1:volsmpl}
\end{figure}

For an additional visualization, Figure~\ref{fig:ability1:depths} (left) plots ordered projection and halfspace depth values for all $100$ observations of the training sample.

\begin{figure}[!t]
	\begin{center}
	\begin{tabular}{cc}
		\quad Projection depth & Halfspace depth \\
		\includegraphics[trim=28.5cm 0cm 35cm 7.5cm,clip=true,scale=.105]{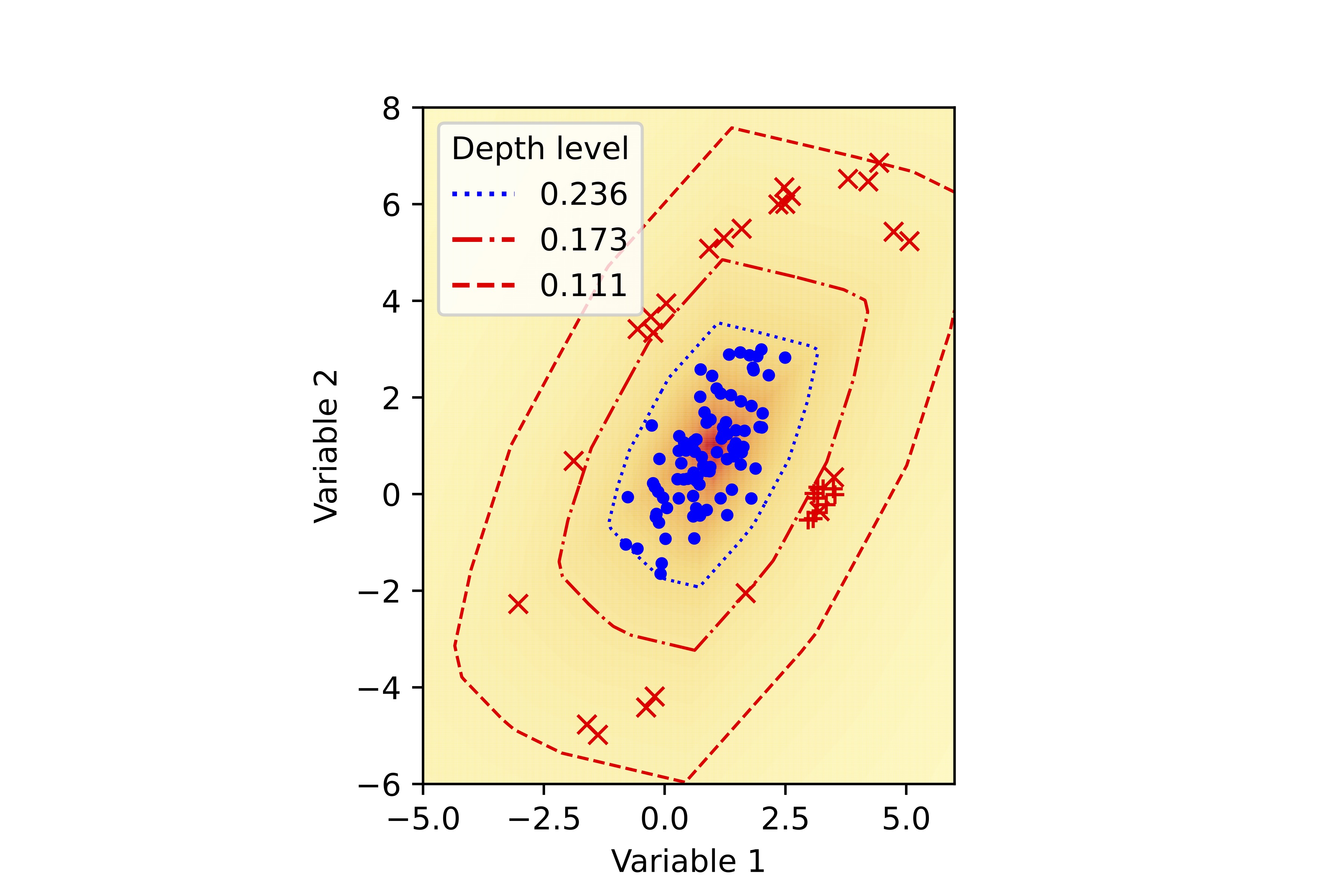} & \includegraphics[trim=27.5cm 0cm 17.5cm 7.5cm,clip=true,scale=.105]{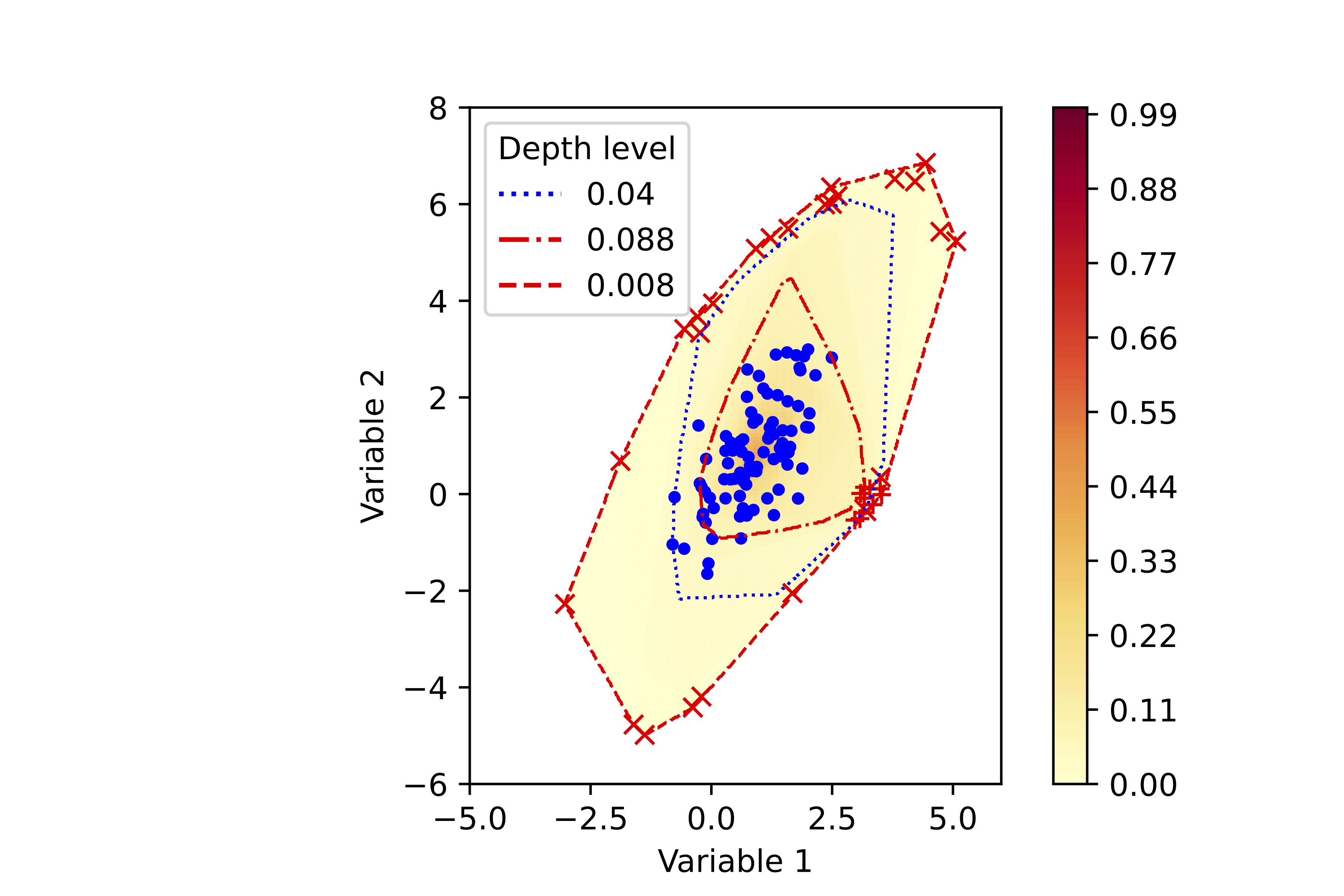}
	\end{tabular}
	\end{center}
	\caption{$90$ observations stemming from bivariate normal distribution (blue dots) contaminated with $10$ clustered (red pluses) and $25$ masking (red crosses) anomalies. Depth contours at three levels: minimal depth of normal observations (blue dotted line), maximal depth of all $35$ anomalies contaminating the training sample (red dash-dotted line), minimal depth of all $35$ anomalies contaminating the training sample (red dashed line). Left: depth values (in color) for simplicial volume depth. Right: depth values (in color) for simplicial depth (with white corresponding to zero). Color scale for both plots is depicted on the right side.}\label{fig:ability2:prjhfsp}
\end{figure}

Let us consider another couple of widely known depths, namely simplicial volume~\eqref{equ:depthOja} and simplicial~\eqref{equ:depthSmpl} depths. For the same training set, similar visualization is depicted in Figure~\ref{fig:ability1:volsmpl}. As before, by construction and due to the immediate vanishing property, simplicial depth fails to detect the group of $10$ anomalies, a task being coped with by simplicial volume depth.

While this example is very typical for anomaly detection, other situations appear. For instance, regard an example where additional (single) anomalies mask the clustered ones. For this, let us add to the previous training sample $25$ observations distributed in a similar manner as normal training data but having Mahalanobis distance (calculated using normal population mean vector and covariance matrix) between minimal and maximal Mahalanobis distance of the $10$ clustered anomalies. As we can see from Figure~\ref{fig:ability2:prjhfsp}, as before projection depth copes with the task correctly retrieving all $35$ anomalies, but this time the halfspace depth also detects majority of them (``thanks'' to masking effect); see additionally Figure~\ref{fig:ability1:depths} (right). Another couple of depths---simplicial volume and simplicial---behave similarly with visualization omitted here for space purposes.

\begin{figure}[!t]
	\begin{center}
	\begin{tabular}{cc}
		\quad Projection depth & Asymmetric projection depth \\
		\includegraphics[trim=6cm 1cm 7cm 4cm,clip=true,scale=.145]{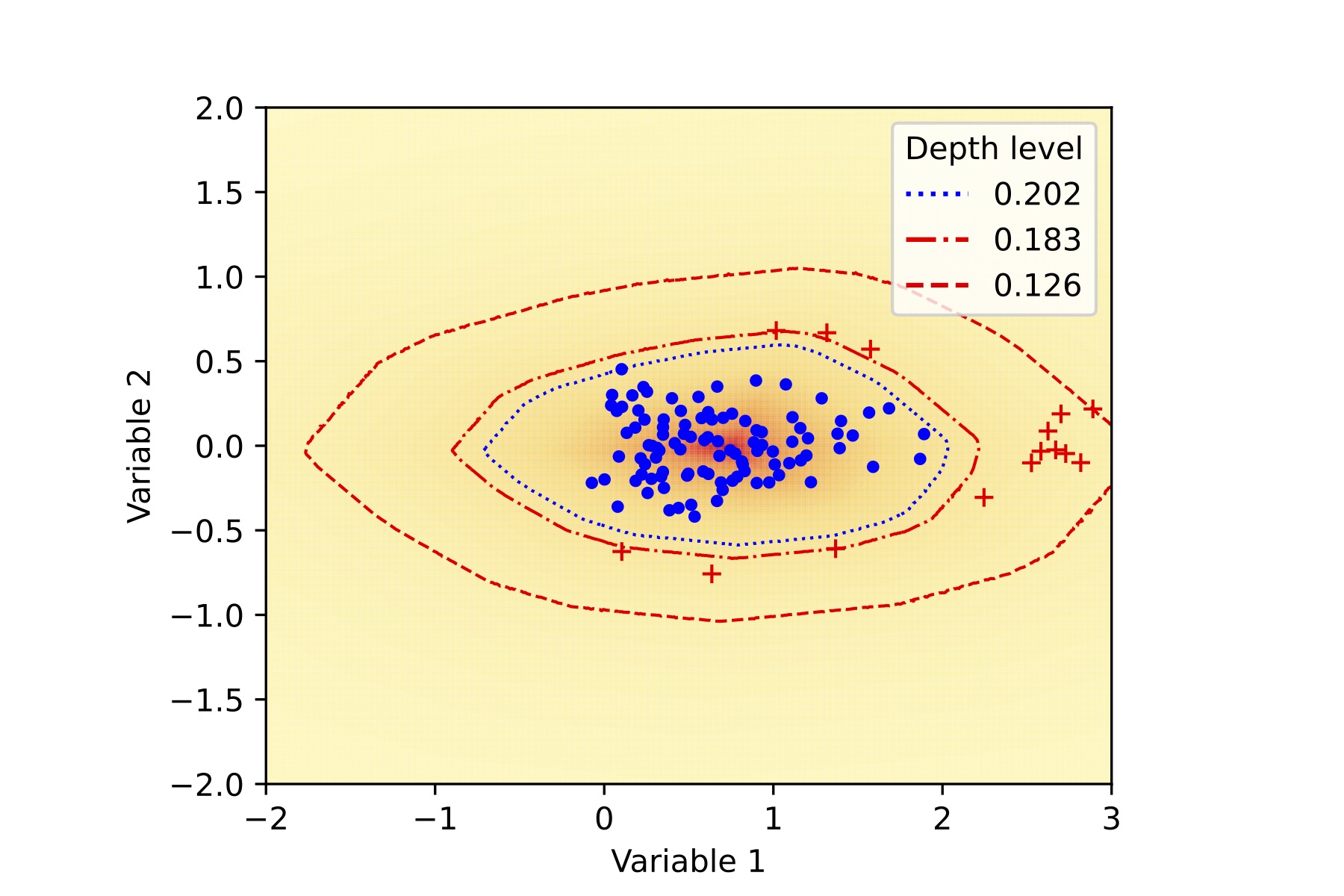} & \includegraphics[trim=1cm 1cm 7.5cm 4cm,clip=true,scale=.145]{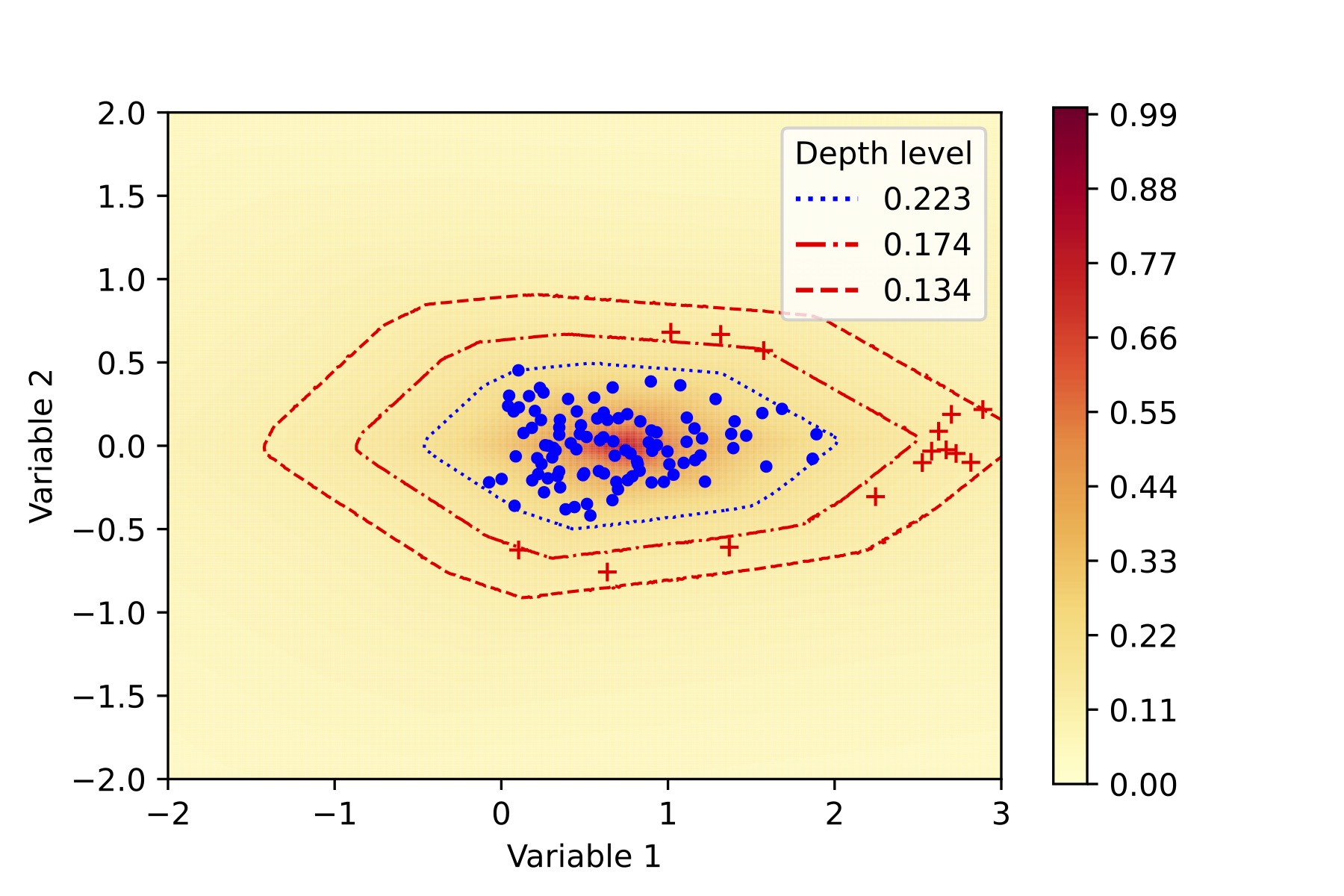} \\
		\includegraphics[trim=0.25cm 0cm 0.75cm 0.5cm,clip=true,scale=.475]{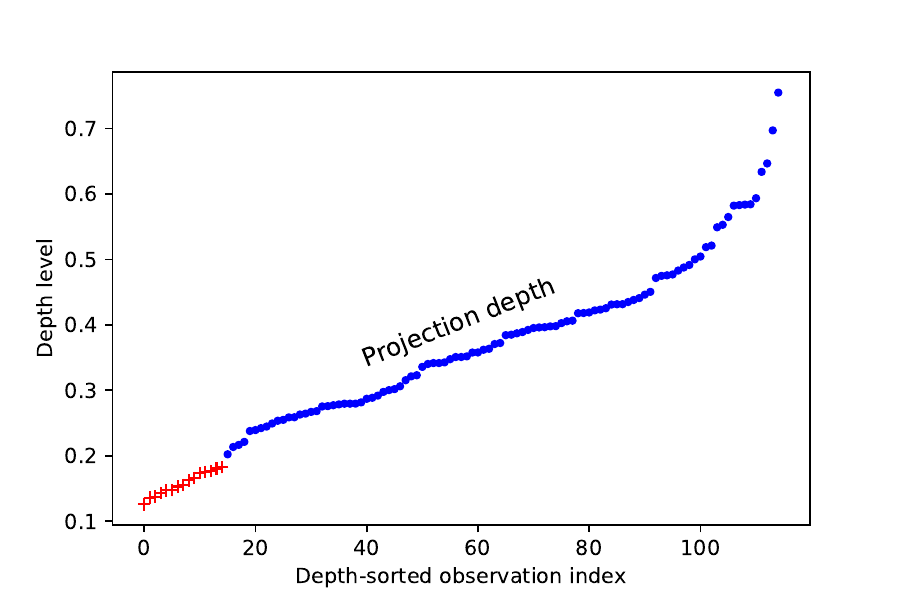} & \includegraphics[trim=0.25cm 0cm 0cm 0.5cm,clip=true,scale=.475]{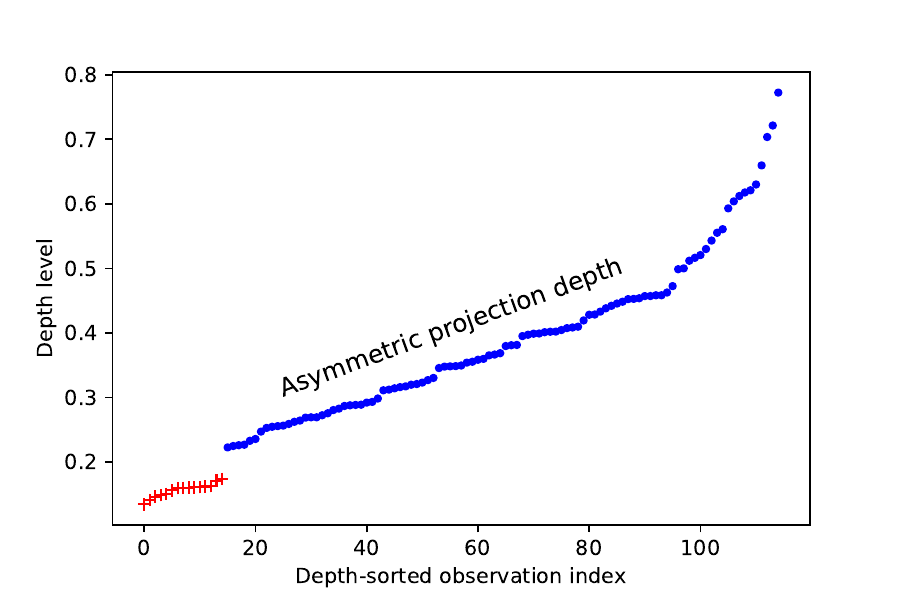}\,\,\,
	\end{tabular}
	\end{center}
	\caption{$100$ observations stemming from skewed bivariate distribution (blue dots) contaminated with $15$ anomalies (red pluses). Depth contours at three levels: minimal depth of normal observations (blue dotted line), maximal depth of $15$ anomalies (red dash-dotted line), minimal depth of $15$ anomalies (red dashed line). Top: depth values (in color) for projection depth (left) and asymmetric projection depth (right); color scale for both plots is depicted on the right side. Bottom: Ordered depth values for projection (left) and asymmetric projection (right) depth.}\label{fig:ability3:prjaprj}
\end{figure}

While projection depth is very suitable for the examples from above, its upper-level sets retain certain degree of symmetry, which can be inadequate for, \textit{e.g.}, skewed data. To get more insights for this latter case, we shall contrast projection depth with its asymmetric version~\eqref{equ:depthAPrj}, in the following setting. First, we generate $100$ (normal) observations from a skewed bivariate distribution with independent marginals being skewed normal with parameter $9$ \citep[abscissa, according to][]{AzzaliniC99} and centered normal with standard deviation $1/4$ (ordinate). Then, we add $15$ anomalies from the same distribution, with probability density $\in[0.01,0.05]$ and positive values for abscissa. The corresponding plots for both depth notions are indicated in Figure~\ref{fig:ability3:prjaprj}, top. One observes, that while in this particular case projection depth still copes with the task, asymmetric projection depth allows to immediately identify the group of anomalies, see Figure~\ref{fig:ability3:prjaprj}, bottom.

\subsection{Choice of the threshold}\label{ssec:threshold}

Together with Section~\ref{ssec:motiv:industry}, this section in a natural way attracts attention to the following practical question: Which value should one choose for the threshold that cuts off anomalies (as observations with small values of depth)?

\begin{figure}[!t]
    \begin{center}
	\begin{tabular}{cc}
            ``Elbow'' threshold & ``Gap'' threshold \\
        \includegraphics[trim=0cm 0cm 1cm 0.75cm, clip=true, scale=.52]{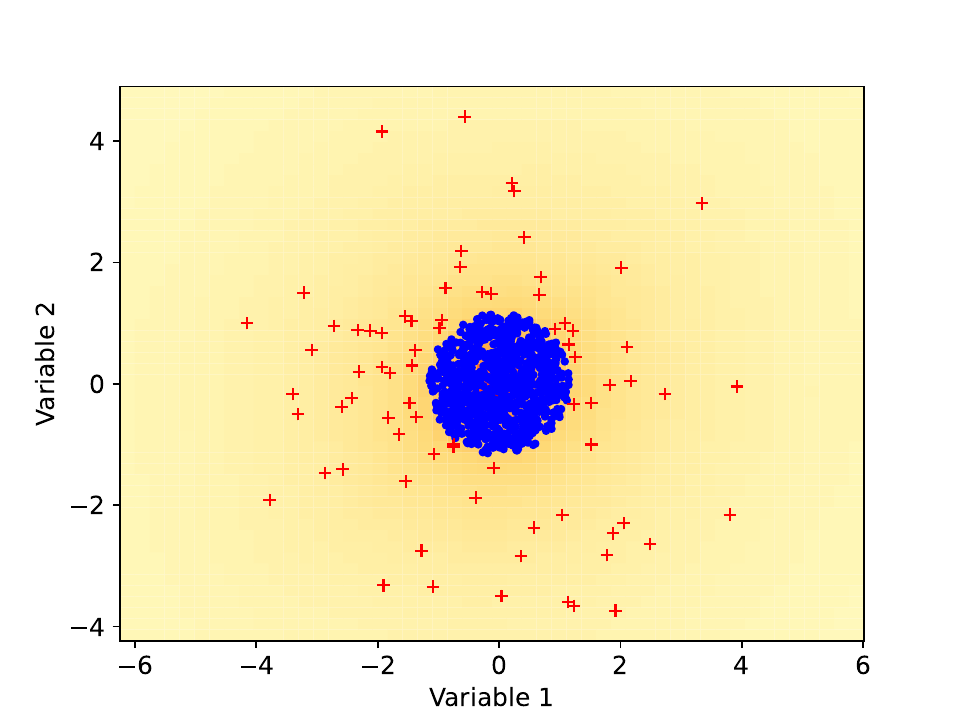} & \includegraphics[trim=0cm 0cm 1cm 0.75cm, clip=true, scale=.52]{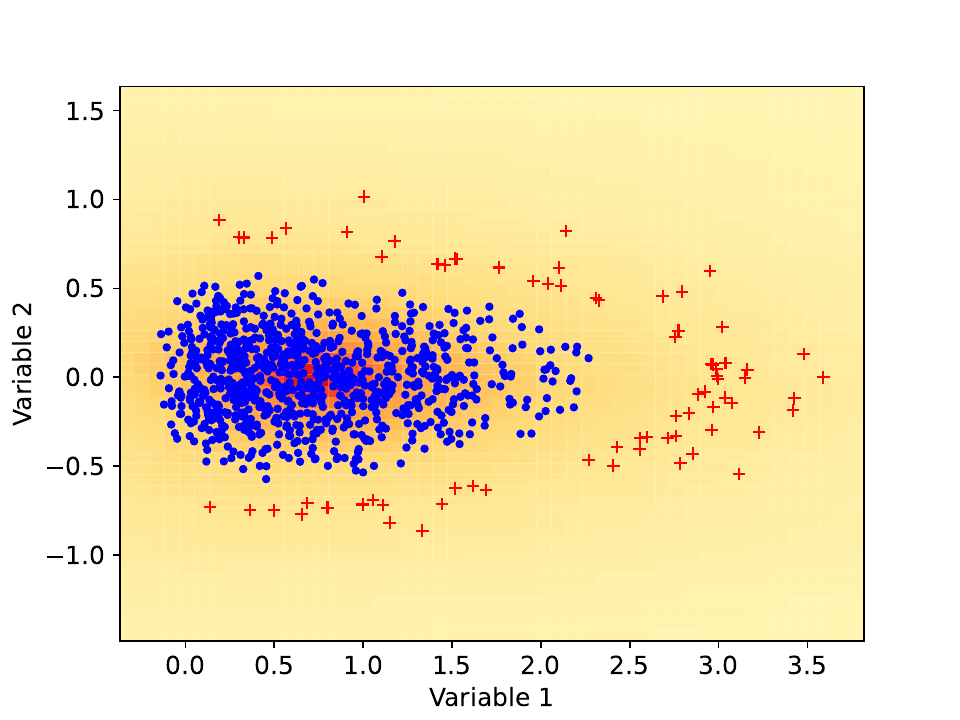} \\
		\includegraphics[trim=0cm 0cm 1cm 0.75cm, clip=true, scale=.52]{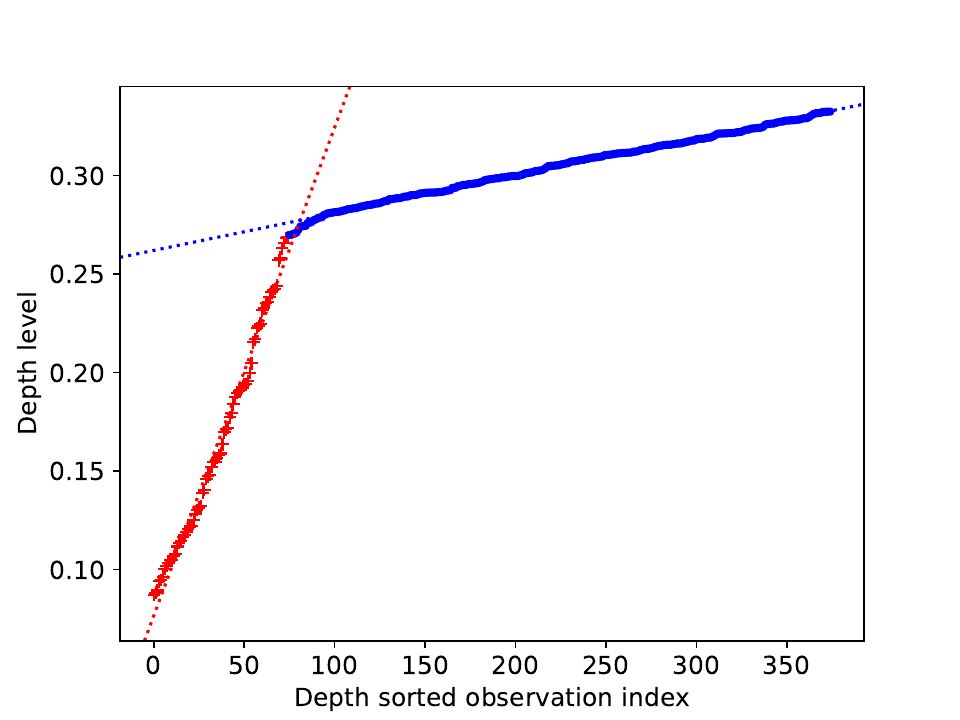} & \includegraphics[trim=0cm 0cm 1cm 0.75cm, clip=true, scale=.52]{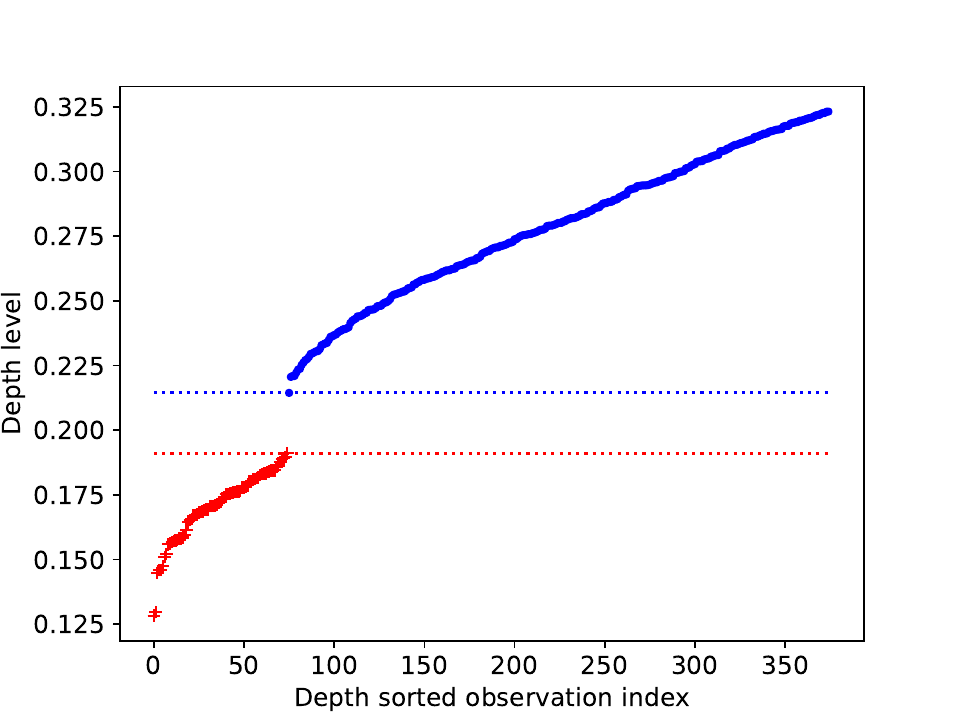}
	\end{tabular}
    \end{center}
    \caption{Simulated data set for an elbow-shaped threshold (left) and for a gap-shaped threshold (right). To ease visualization, depth level is zoomed on the lower $30\%$ values.}\label{fig:threshold}
\end{figure}

In view of generality and complexity of the task of unsupervised anomaly detection, it would be apparently too hopeful to expect a universal answer. One practical way is to use the so-called ``elbow rule'', \textit{i.e.}, to place threshold where the slope of the (locally) straight line following plotted---in decreasing order---depth values suddenly decreases. This works if plotted in decreasing (or increasing) order depth values can be well interpolated by two connected straight lines such as in left Figure~\ref{fig:threshold} (the threshold is then set where the two lines connect), but is much more ambiguous otherwise. Another practical way is to place the threshold in the ``gap'' between ordered depth values, as this can be observed in Figures~\ref{fig:ability1:depths} (both left and right) for projection depth and Figure~\ref{fig:ability3:prjaprj} (right) for asymmetric projection depth or Figure~\ref{fig:threshold} (right) for an explicit example; as before application of this rule becomes ambiguous if such a gap on the depth graph does not exist.

Furthermore, in practice the anomaly-detecting threshold is rarely fixed (forever) in advance and is periodically altered, either due to distribution change or (enterprise) policy evolution. Guided by such a policy, one naturally attempts to (always) detect all anomalies, which can optimistically be the case, but more often one has to admit that part of anomalies will remain unnoticed. In this latter case it is important to deliver a reasonable score reflecting a degree of abnormality---advocated as data depth in this article. Such a score shall further allow for adjusting the threshold depending on the application-related trade-off, \textit{e.g.}, between undetected anomalies (false negatives) and normal data marked as anomalies (false positives); there are of course other formulations of this trade-off using different quantities.

It is noteworthy that getting a deeper insight in the nature of anomalies, \textit{e.g.}, by explaining those (presumably) detected in the training sample (see Section~\ref{ssec:adv:explain} below for an example) is beneficial for threshold definition as well.

For the reasons described above, the approach used in benchmark studies of Sections~\ref{sec:advantages} and~\ref{sec:comptract} shall be based on evaluation of the obtained ordering; see also~\eqref{equ:critboxplot}.


\section{Advantages}\label{sec:advantages}

As announced in the introduction, the main goal of this section is to illustrate advantages of the data depth when performing anomaly detection. In order to do this, its subsections shall include comparisons with the most used methods for anomaly detection existing in the literature. It is important to stress, that our goal here is not to state that data depth performs better---in frame of anomaly detection---than the mentioned methods neither to provide a comprehensive comparison, but rather to identify potential cases where its deployment can be advantageous. 

In Section~\ref{ssec:adv:ae} we start with illustrating robustness of the depth-based methodology comparing with widely used anomaly detection neural network---auto-encoder. Further, in Section~\ref{ssec:adv:extrap} depth is contrasted with three major methods for anomaly detection: isolation forest, local outlier factor, and one-class support vector machine. Finally, Section~\ref{ssec:adv:explain} shall highlight explainability capacities of the depth-based approach---a highly demanded feature today lacking for neural networks.

Anomaly detection rule~\eqref{equ:projdepth} based on projection depth~\eqref{equ:depthPrj} shall be used throughout this section.

\subsection{Robustness}\label{ssec:adv:ae}

\emph{Autoencoder} is a widely used contemporary method for anomaly detection, applied to various types of data. The idea is to use bottle-neck when joining sequentially neuronal encoder and decoder, and detect anomalies by high reconstruction loss. Being a neural network, it inherits all the advantages of scalability and flexibility from artificial-neuron-based architectures, demanding meticulous tuning and (especially for complex data sets) computational resources though. Complexity of the autoencoder model (which can be quantified by its exact architecture as well as by the number of parameters to estimate, with the last number sometimes reaching millions) impedes human-understandable interpretability of its output. Elaborating on this explainability issue later in Section~\ref{ssec:adv:explain}, let us discuss autoencoder's robustness contrasting it with data depth.

\emph{Deep support vector data description} (deep SVDD) acts similarly to the spherical one-class support vector machine~\citep{ScholkopfPSTSW01}, but (like autoencoder) improves the anomaly detection performance using neural-network-based representation instead of RKHS~\citep{Tax2004}. With the method being widely used in practice and inheriting advantages from both statistical and machine learning sides, we include it in the comparison.


In the unsupervised setting, when no feedback is given about abnormality of the observations of the training set, autoencoder detects anomalies assuming that majority of the data are normal and deviations of anomalies are not critically large. This is due to the fact that, when being trained using stochastic gradient descent algorithm, derivative of the loss function is required. This smoothness condition impedes autoencoder from being robust, since larger losses---in particular with substantial portions of anomalies---``distract'' the model (on an average) during training. This same assumption of majority of data being normal also holds for Deep SVDD, and in presence of higher fractions of anomalies in the training data can lead to a poor embedding of normal data in the neural network part and consequently may result in a bad hypersphere minimisation.

A robust depth function (\textit{e.g.} projection or halfspace depth), on the other hand, can include elements of indicator nature (corresponding to the so-called $0$-$1$-loss) in the definition, which are not influenced by the amplitude of anomalies. Such indicator-containing functions do not posses derivative everywhere (and even have steps/breaks), which makes them unoptimizable for neural network architectures. Such an approach rather shifts the problem from the definition part (\textit{e.g.}, surrogate loss) to the numerical/computational stage; solutions to this problem has been recently proposed for a number of data depth functions in larger dimensions~\citep[see, \textit{e.g.},][]{DyckerhoffMN21}. (In the present article, computation of data depth covers spaces up to $\mathbb{R}^{50}$.) We illustrate this difference on a short simulation study right below.

Consider a training data set $\bmX_{tr}\subset\mathbb{R}^d$ consisting of $n=700$ observations simulated in the following way. First, subset of $\lfloor n (1-\varepsilon) \rfloor$ (normal) points $\bmY_{tr}=\{\bmy_1,\bmy_2,...,\bmy_{\lfloor n (1-\varepsilon)\rfloor}\}$ is drawn from the normal distribution in $\mathbb{R}^d$ ($\mathcal{N}(\bmi_d,\bmI_{d\times d}),$ where $\bmi_d=(1,1,...,1)^\top$ is the $d$-vector of $1$s and $\bmI_{d\times d}$ is the $d\times d$ matrix of $0$s with $1$s on the main diagonal only). Then, second subset $\bmZ_{tr}$ of $\lceil n \varepsilon\rceil$ (abnormal) points is drawn from the conditional Cauchy distribution $Z\,|\,\|Z\|>1.5\max(\|\bmy_1\|,\|\bmy_2\|,...,\|\bmy_{\lfloor n (1-\varepsilon)\rfloor}\|)$, where $Z$ is the $d$-variate random vector stemming from elliptical Cauchy distribution. (Elliptical distribution is a generalization of the multivariate normal distribution with density contours being ellipsoids to further univariate laws then the Gaussian; the reader is referred to, \textit{e.g.},~\cite{FangKN90} for more details.) Third, $\bmX_{tr}=\bmY_{tr}\cup\bmZ_{tr}$, followed by (numerical) random shuffling of the elements of $\bmX_{tr}$. Likewise, the same procedure is used to generate testing data $\bmX_{te}$, with $n=300$.

The depth-based anomaly scoring rule of type~\eqref{equ:projdepth} is then applied to each observation of the testing data, based on projection depth. The smallest threshold is chosen to correctly detect all anomalies in $\bmX_{te}$: $t_{\text{prj},\bmX_{te}}=\max_{\bmz\in\bmZ_{te}}D^{\text{prj}}(\bmz|\bmX_{tr})+\epsilon$ for some infinitesimal positive $\epsilon$. It is shown this threshold is built on new observations $\bmX_{te}$ but using $\bmX_{tr}$ as a reference to emulate the standard machine learning framework (training followed by testing). The following quantity is then used for comparison:
\begin{equation}\label{equ:critboxplot}
	p(\bmX_{te}) = \frac{\sum_{\bmz\in\bmZ_{te}} g_{\text{prj}(\bmz|\bmX_{tr})}}{\sum_{\bmz\in\bmX_{te}} g_{\text{prj}(\bmz|\bmX_{tr})}}\,,
\end{equation}
which reflects the (largest) portion of the anomalies in the---ordered by depth---part of the testing sample identified as anomalies such that all anomalies are correctly detected.

\begin{figure}[!t]
	\begin{center}
	\begin{tabular}{cc}
		$d = 10$ & $d = 20$ \\
		\includegraphics[trim=0.5cm 0cm 1cm 1cm,clip=true,scale=.535]{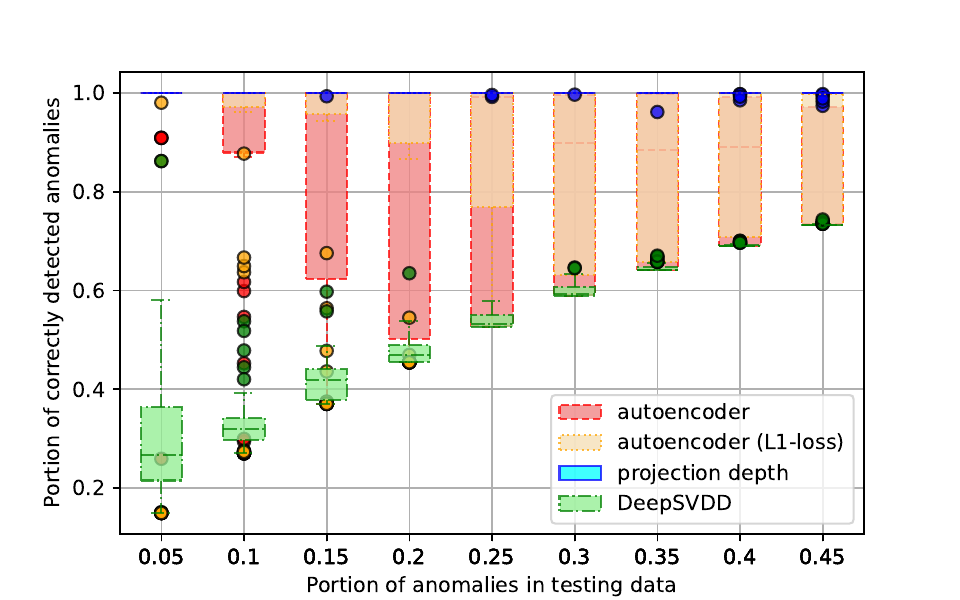} & \includegraphics[trim=0.5cm 0cm 1cm 1cm,clip=true,scale=.535]{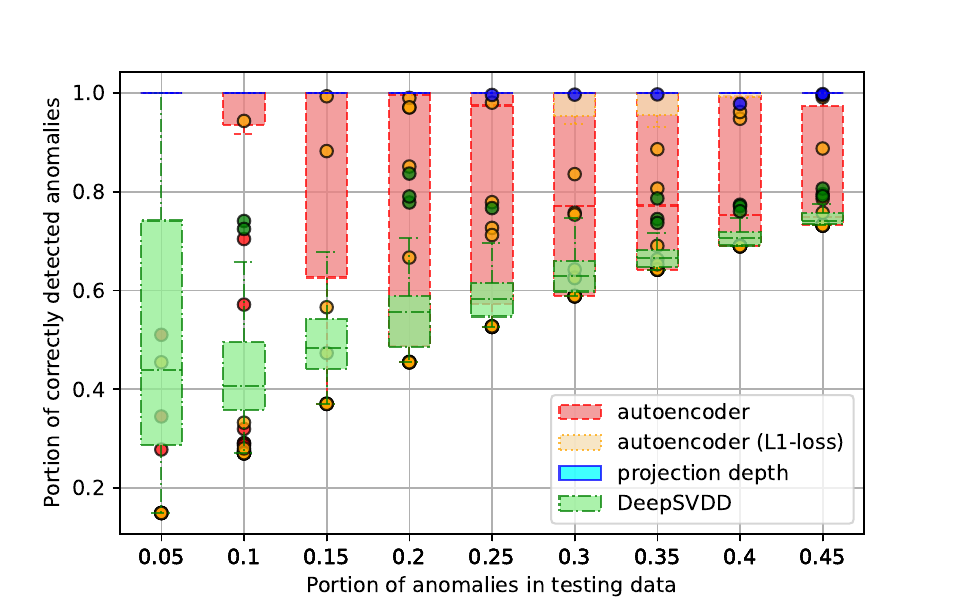}
	\end{tabular}
	\end{center}
	\caption{Boxplots of portion of correctly detected anomalies of the testing data set when setting the threshold to detect all anomalies, over $50$ random draws. Normal data stems from the multivariate Gaussian distribution in $\mathbb{R}^d$ ($d=10$ on the left and $d=20$ on the right) while anomalies are drawn from elliptical Cauchy distribution and having distance $>1.5$ of the most distant normal observation from the origin. The three methods are: autoencoder with quadratic loss (red, dashed), autoencoder with $L_1$ loss (orange, dotted), Deep SVDD (green, dashed-dotted) and the projection depth (blue, solid).}\label{fig:robustae}
\end{figure}

With autoencoder, the anomaly scoring rule is based on (quadratic) reconstruction error: observations with higher (\textit{i.e.}, beyond a threshold) reconstruction error are identified as anomalies. As well as in~\eqref{equ:critboxplot}, the threshold is chosen to be (slightly smaller than) the highest value to recognize correctly all anomalies.
The autoencoder is trained on $\bmX_{tr}$ with both quadratic and $L_1$ losses, but only quadratic loss is used to detect anomalies. Boxplots of $p$ from~\eqref{equ:critboxplot} over $50$ independent draws of $\bmX_{te}$ for both autoencoders, as well as for the projection depth, are indicated in Figure~\ref{fig:robustae}, where $\varepsilon=0.05,0.1,...,0.45$ and $d=10,20$ are tried. (Projection depth was approximated using Nelder-Mead algorithm as in~\cite{DyckerhoffMN21} with $100$ directions for $d=10$ and $200$ directions for $d=20$. The autoencoder contains three hidden layers each having $5-2-5$ (for $d=10$) neurons ($20-10-5-10-20$ for $d=20$, respectively) and is trained $100$ epochs with $10$-observations per minibatch, using stochastic gradient descent algorithm with learning rate $0.005$. Python libraries \texttt{data-depth} and \texttt{PyTorch} were used, respectively.)

From Figure~\ref{fig:robustae}, one can clearly observe that while depth-based anomaly detection perfectly copes with the task, the autoencoder fails in most of the cases if portion of anomalies in the training set exceeds $10\%$ and with improving but becoming more concentrated error if this portion grows. Further, higher non-stability of training of autoencoder is observed, while introducing (robust) $L_1$ loss does not improve the output much. It is important to notice that the autoencoder's architecture and training schedule/loss used here are typical and chosen for illustration purposes; when robustness is a potential issue more sophisticated architectures as well as losses and training schedules could improve the results.

\subsection{Extrapolation}\label{ssec:adv:extrap}

While fundamental results of the empirical risk minimization theory request the distributions of the training and the test data be identical, anomaly detection preserves additional difficulties since operational abnormal patterns' behaviour can differ (substantially) from anomalies present in the training data. To tackle this issue, employed anomaly detection method at hand should be able to extrapolate the knowledge about normal data to ``protect'' it from (preferably) all possible occurrences of future anomalies.

In this subsection we shall pay more attention to the operational nature of anomaly detection, \textit{i.e.}, the presence of decision function (or a rule), training data set, and new data (often called test set in study setting). For illustration purposes, we contrast data depth with \textit{local outlier factor}~\citep[][LOF]{BreunigKNS00}, \textit{one-class support vector machine}~\citep[][OC-SVM]{ScholkopfPSTSW01}, and \textit{isolation forest}~\citep[][IF]{LiuTZ08} on the same training and test data sets, which we construct as follows.

$100$ observations $\in\mathbb{R}^2$ constitute (viewable) \textit{training data set}. $90$ of them---normal data---follow uncorrelated normal distribution with mean $(1/2, 1/2)^\top$ and equal standard deviations $1/4$ for both variables. $10$ anomalies are drawn from uncorrelated normal distribution with mean $(-3/4, 1/2)^\top$ and both standard deviations equal $1/10$. For the \textit{testing set} (consisting of $300$ observations), $250$ observations are generated from the distribution of normal data, $25$ from the distribution of anomalies, and $25$ from uncorrelated normal distribution with mean $(7/4, 1/2)^\top$ and both standard deviations equal to $1/10$.

Parameters of the anomaly detection methods were chosen as follows: For LOF, number of neighbors is set to $25$; OC-SVM is used with Gaussian kernel with parameter $\gamma=0.1$ and regularization constant $\nu=0.1$ (which after has illustrated best results); IF is constituted of $500$ isolation trees; depth-based rule~\eqref{equ:projdepth} is used with projection depth approximated with the spherical Nelder-Mead algorithm using $500$ directions. (After multiple tries, values delivering best results on these data have been chosen; eventually the methods exhibit their regular behavior.) \texttt{Python} libraries \texttt{scikit-learn}~\citep{scikitlearn} and \texttt{data-depth} were used for implementation.

\begin{figure}[!t]
	\begin{center}
		\includegraphics[trim=2cm 2cm 12.5cm 9cm,clip=true,scale=.07]{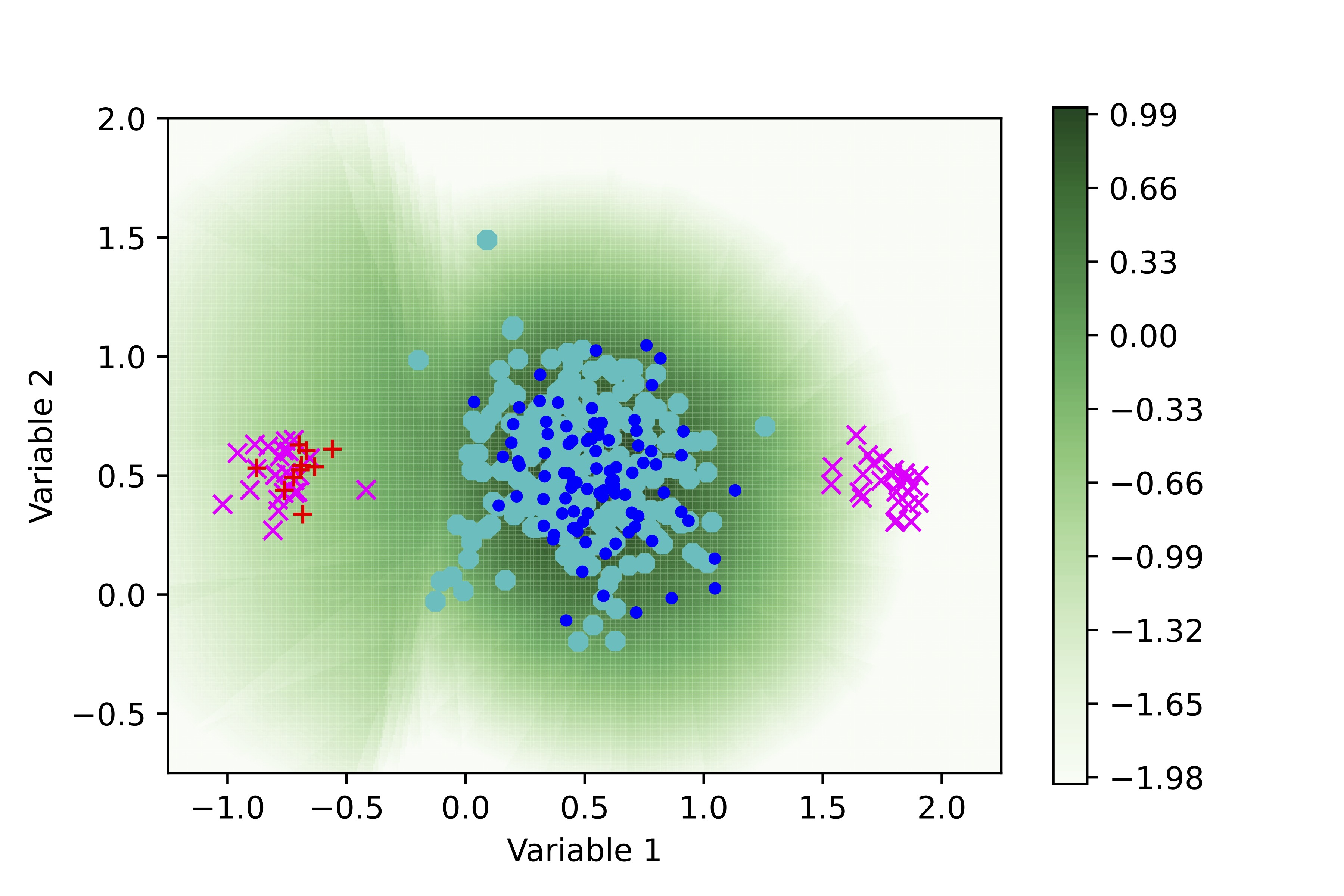}\quad\includegraphics[trim=0cm 0.1cm 1.25cm 1cm,clip=true,scale=.5675]{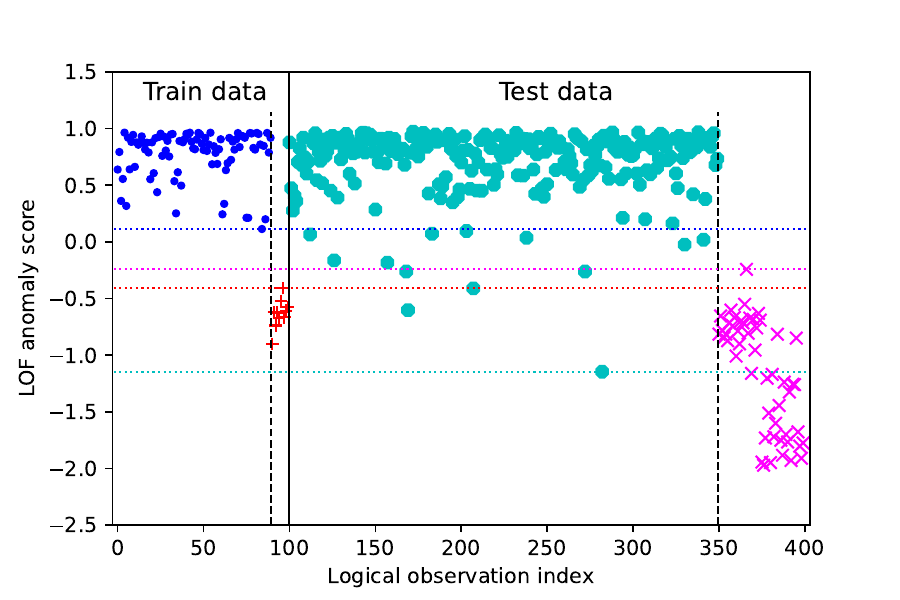}
	\end{center}
	\caption{Anomaly detection using LOF, for the data from Section~\ref{ssec:adv:extrap}. Left: Plot of training (blue dots for normal observations and red pluses for anomalies) and testing (bigger cyan dots for normal observations and magenta crosses for anomalies) sets and anomaly score for the entire space $\mathbb{R}^2$, with scale. Right: Using same markers, for both training and testing sets, anomaly score (on ordinate) by LOF. Separated by vertical lines from left to right are normal observations (indices on abscissa $1$-$90$) and anomalies ($91$-$100$) of the training set, normal observations ($101$-$350$) and anomalies ($351$-$400$) of the test set.}\label{fig:extra:lof}
\end{figure}

Figure~\ref{fig:extra:lof} illustrates anomaly detection of both training and test sets using LOF. In general, LOF copes with the task to certain degree, where only very little normal observations of the testing set (which are further from the center) have same score as anomalies. Moreover, better choice of number of nearest neighbors could even improve the results. The only minor disadvantage is that anomalies on the right side of the normal data (\textit{i.e.} where there were no anomalies in the training set) have different score from those on the left side, which can create a false impression that they are more abnormal, but indeed normal data were contaminated. As we shall see in Appendix~\ref{app:extrap:10:20}, this behaviour is as well preserved with growing dimension, while the accuracy of a neighbour-based method is expected to deteriorate.
Further, the number of nearest neighbors should be chosen, which can be a difficult task in the non-supervised context, and may not always provide satisfactory results if the data's density varies.

\begin{figure}[!t]
	\begin{center}
		\includegraphics[trim=0cm 0cm 1.25cm 1cm,clip=true,scale=.565]{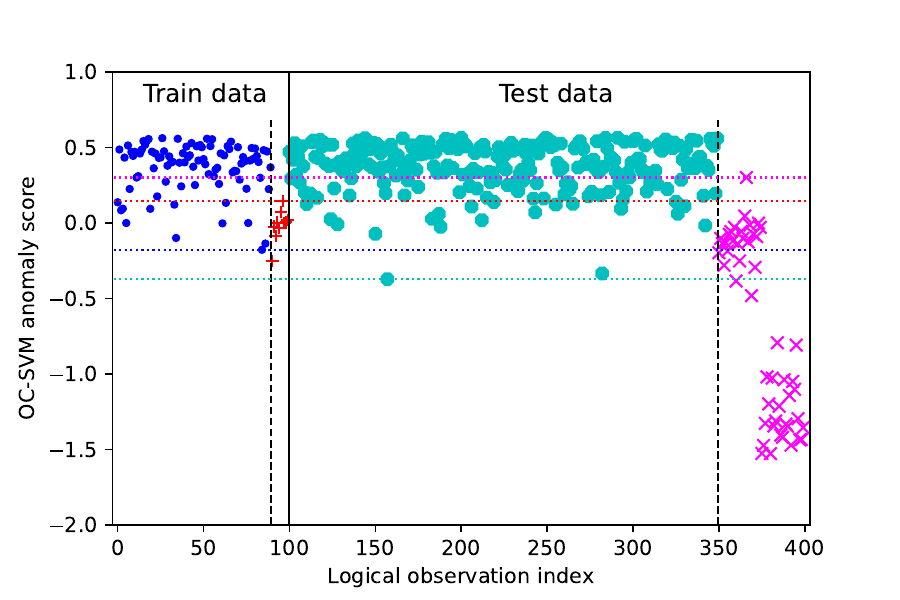}\quad\includegraphics[trim=0cm 0cm 1.25cm 1cm,clip=true,scale=.565]{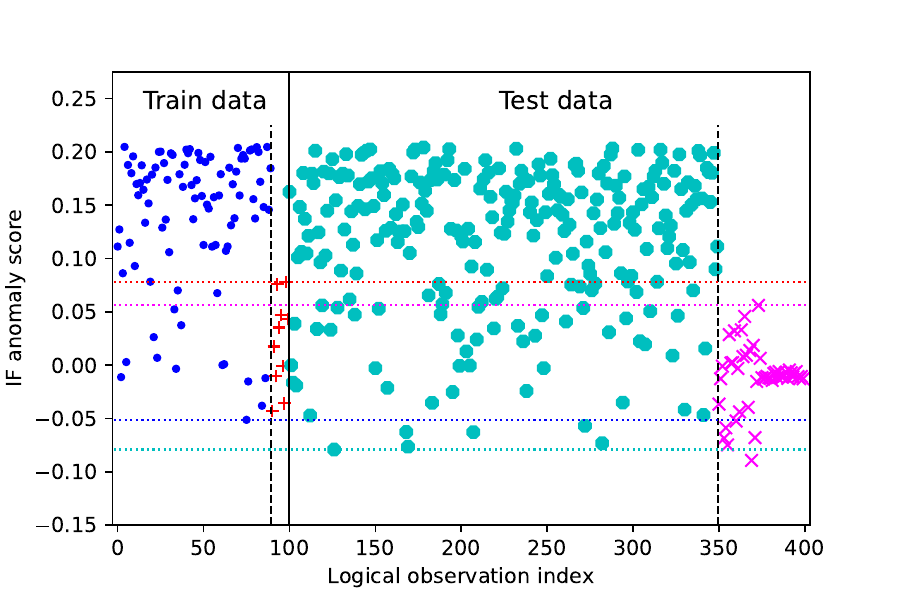}
	\end{center}
	\caption{For both training and testing sets, anomaly score (on ordinate) by OC-SVM (left) and IF (right). Separated by vertical lines from left to right are  normal observations (indices on abscissa $1$-$90$, blue dots) and anomalies ($91$-$100$, red pluses) of the training set, normal observations ($101$-$350$, bigger cyan dots) and anomalies ($351$-$400$, magenta crosses) of the test set.}\label{fig:extra:ocsvmif}
\end{figure}

Figure~\ref{fig:extra:ocsvmif} (left) indicates the anomaly score of OC-SVM. Being initially designed for estimation of support, OC-SVM fails to detect anomalies (even though assigning them rather low score, proper for non-central data) where they were already present in the training set, and detects only newly introduced anomalies. Better tuning of the kernel bandwidth does not seem to substantially change the plot. It is noteworthy that for a training set without (or with only several) anomalies OC-SVM would perform better, while also coping with the curse of dimension.

Anomaly score of IF is depicted in Figure~\ref{fig:extra:ocsvmif} (right). It does not detect any of the anomalies, with the reason being their location on the level of the center of normal data in one of the coordinates (ordinate). This happens while IF treats the dimensions one-by-one, and becomes less important with growing dimension since the volume of such areas decreases. \cite{HaririKB21} suggest a modification of IF (called extended isolation forest), which using random directions suppresses this effect. It is of course important to mention, that scores of anomalies are nevertheless low, similar to those of non-central normal data.

\begin{figure}[!t]
	\begin{center}
		\includegraphics[trim=2cm 2cm 12.5cm 10cm,clip=true,scale=.07]{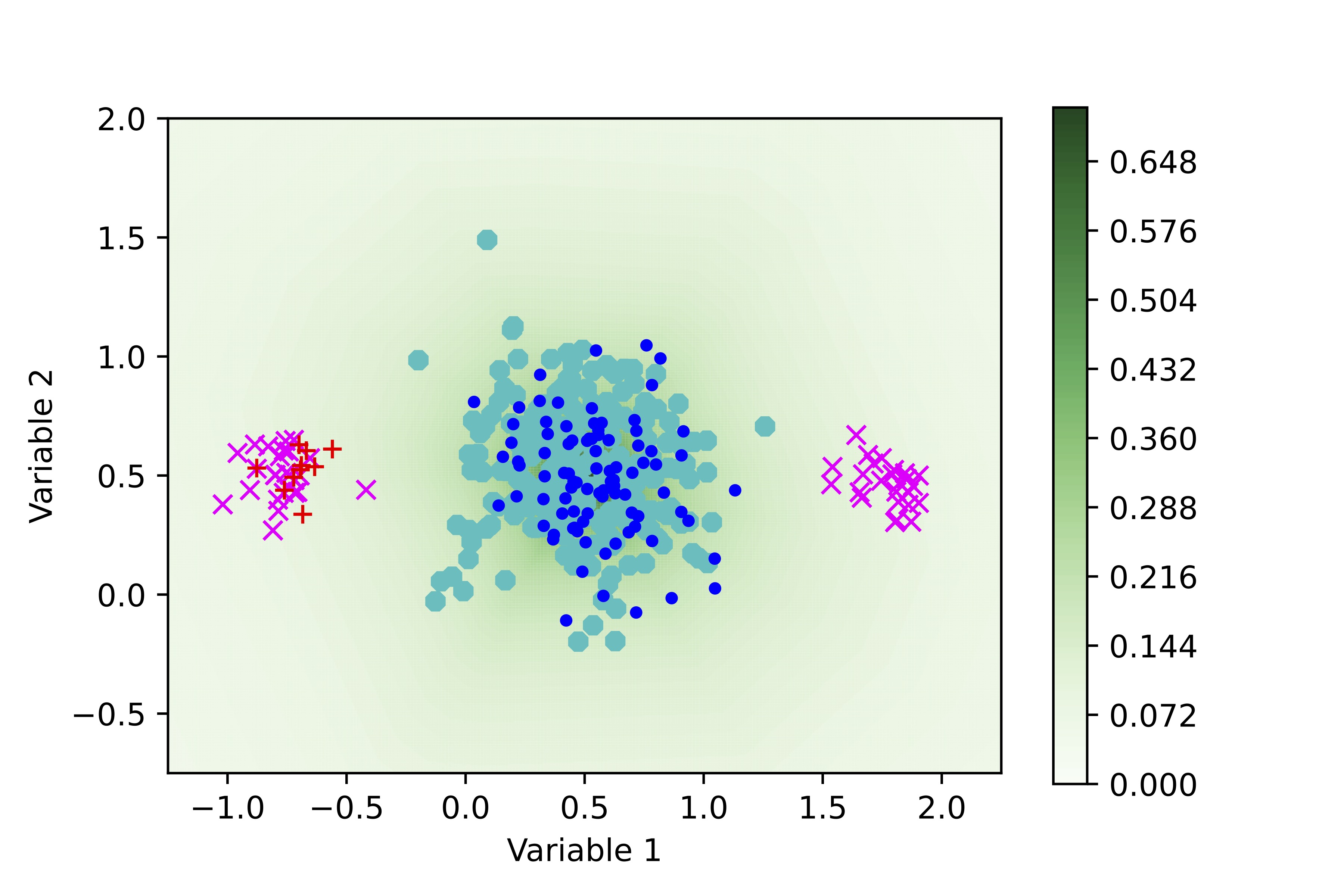}\quad\includegraphics[trim=.5cm 0cm 1.25cm 1cm,clip=true,scale=.5675]{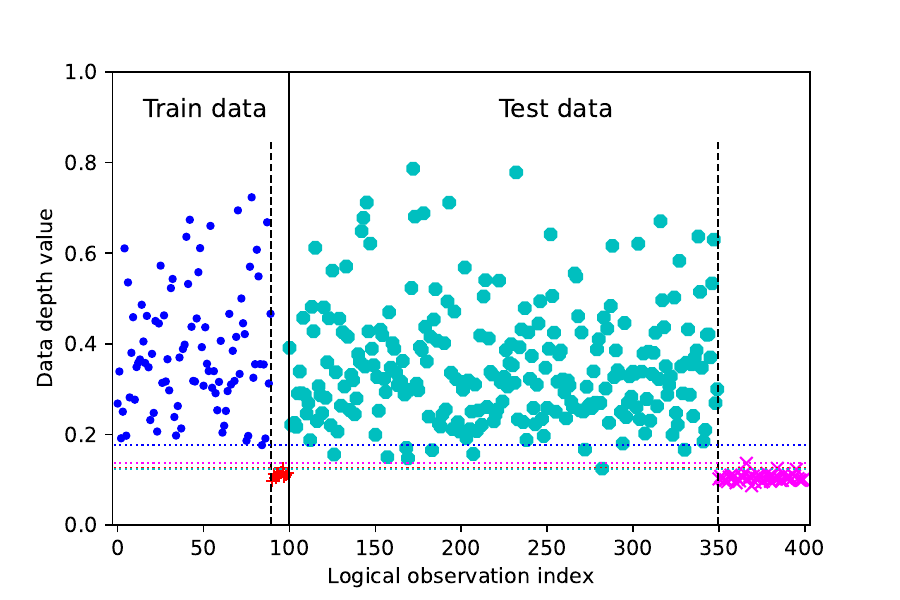}
	\end{center}
	\caption{Anomaly detection using data depth, for the data from Section~\ref{ssec:adv:extrap}. Left: Plot of training (blue dots for normal observations and red pluses for anomalies) and testing (bigger cyan dots for normal observations and magenta crosses for anomalies) sets and anomaly score for the entire space $\mathbb{R}^2$, with scale. Right: Using same markers, for both training and testing sets, anomaly score (on ordinate) by data depth. Separated by vertical lines from left to right are normal observations (indices on abscissa $1$-$90$) and anomalies ($91$-$100$) of the training set, normal observations ($101$-$350$) and anomalies ($351$-$400$) of the test set.}\label{fig:extra:depth}
\end{figure}

In this case, data depth (see Figure~\ref{fig:extra:depth}) well copes with the task and detects anomalies in both training and testing set, including the group of newly introduced anomalies on the right of the normal data. Due to its robustness, the depth of both groups of anomalies is very similar reflecting their equal degree of abnormality. Similar experiments with $n=1000$ points in dimensions 10 and 20 are available in Appendix~\ref{app:extrap:10:20}.

\subsection{Explanation of anomalies}\label{ssec:adv:explain}

Not only detecting anomalies, but also providing explanation about their nature is a highly demanded contemporary topic treatable by data depth. Let us take a deeper look at the first example of Section~\ref{sec:suitability}; when using projection depth, visualizations are provided in Figures~\ref{fig:ability1:prjhfsp} (left) and~\ref{fig:ability1:depths} (left). In this subsection, only the training data $\bmX_{tr}$ will be used with the goal being to explain anomalies, while the same idea can be readily applied to test data $\bmX_{te}$. When applying rule~\eqref{equ:projdepth}, choosing a proper threshold $t_{\text{prj},\bmX_{tr}}$ (\textit{e.g.} $0.175$) does not seem to be a difficult task in this particular case: even when imagining that abnormal observations do not differ in marker, a characteristic jump in (low) depth values provides a good indication.

Further, again sticking to projection depth, explanation of anomalies can be obtained when studying them revealing directions. \textit{E.g.}, for the observation with the smallest depth ($\approx 0.131$), direction minimizing projection depth has coordinates $(0.863, -0.505)^\top$, which is very close to the second principal vector (of normal data) which equals $(0.872, -0.489)^\top$ as we have generated the anomalies. These vector's constituents may indicate which variables and to which degree are ``responsible'' for the abnormality (of this most abnormal) observation.

\begin{figure}[!t]
    \begin{center}
        \includegraphics[trim=0cm 0cm 1cm 1cm, clip=true, scale=.56]{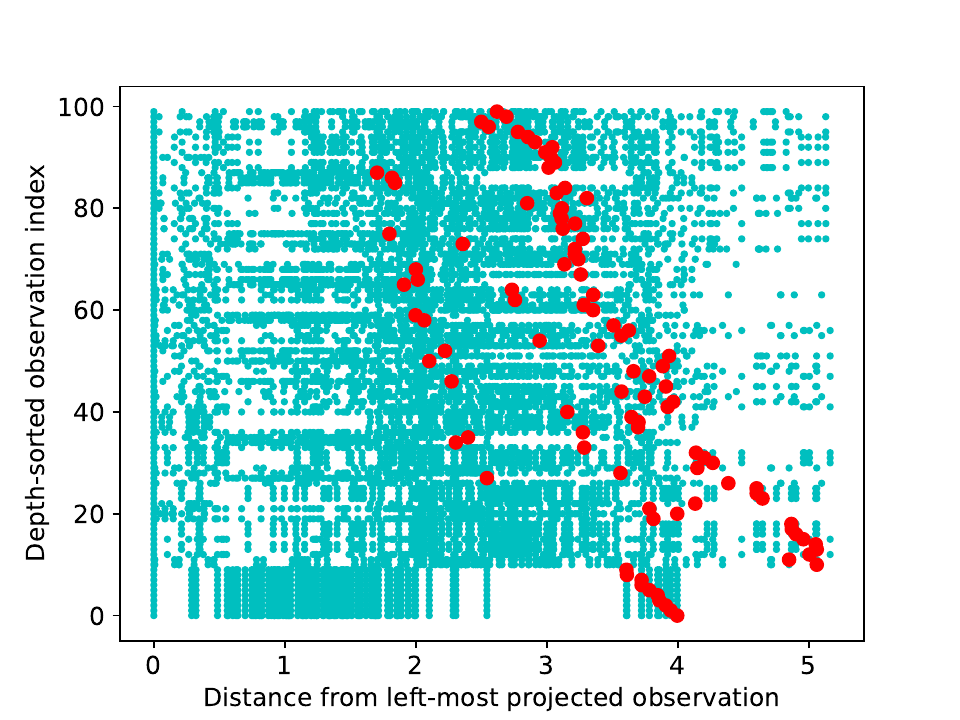}\,\includegraphics[trim=1cm 0cm 1cm 1cm, clip=true, scale=.56]{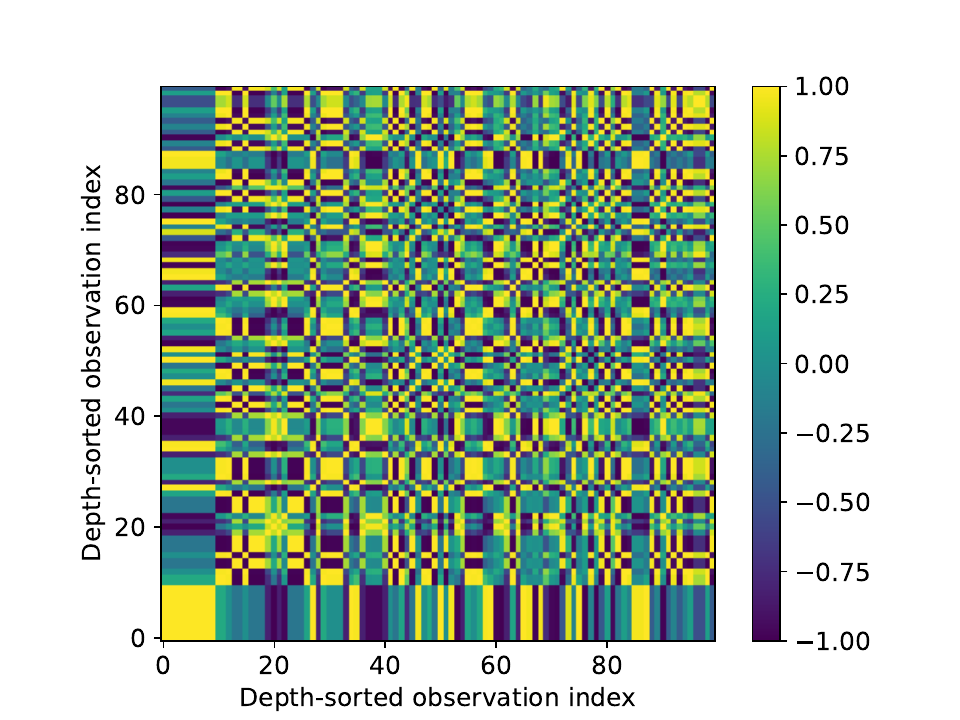}
    \end{center}
    \caption{Left: Projections of data set's observations (cyan dots) from first example of Section~\ref{sec:suitability} on directions minimizing (projection) depth (optimal directions) for each of the observations of the data set; value of left-most projection is subtracted from each projection. Projections of the observations for which the depth was computed are red points. The projections are ordered by increasing depth value of the observations for which the depth was computed. Right: Scalar products between optimal directions (for projection depth) for each observation (ordered by their depth values) of the data set from first example of Section~\ref{sec:suitability}.}\label{fig:explanation:prj}
\end{figure}

Let us now take a look at the two following visualizations of these directions for the entire data set. First, for each $\bmx_i\in\bmX_{tr}$, let $\mathcal{S}^{d-1}\ni\bmu^*_{\bmx_i}=\argmin_{\bmu\in\mathcal{S}^{d-1}} D^{\text{prj}}(\bmx_i|\bmX_{tr})$ be the direction that minimizes~\eqref{equ:depthPrj} (we will call it \textit{optimal direction} for $\bmx_i$). Further, consider the sequence of observations' projections on this optimal---for observation $\bmx_i$---direction:
\begin{equation*}
	\left({\bmu^*_{\bmx_i}}^\top\bmx_{1_{\bmu^*_{\bmx_i}}} - m_{\bmu^*_{\bmx_i}},{\bmu^*_{\bmx_i}}^\top\bmx_{2_{\bmu^*_{\bmx_i}}} - m_{\bmu^*_{\bmx_i}},...,{\bmu^*_{\bmx_i}}^\top\bmx_{n_{\bmu^*_{\bmx_i}}} - m_{\bmu^*_{\bmx_i}}\right)\,,
\end{equation*}
where $m_{\bmu^*_{\bmx_i}}=\min_{j\in\{1,...,n\}}{\bmu^*_{\bmx_i}}^\top\bmx_{j_{\bmu^*_{\bmx_i}}}$ and ${\bmu^*_{\bmx_i}}^\top\bmx_{1_{\bmu^*_{\bmx_i}}} \le {\bmu^*_{\bmx_i}}^\top\bmx_{2_{\bmu^*_{\bmx_i}}} \le ... \le {\bmu^*_{\bmx_i}}^\top\bmx_{n_{\bmu^*_{\bmx_i}}}$. For each observation of the same data set, sorted in depth increasing order (on the ordinate), these sequences are plotted in Figure~\ref{fig:explanation:prj} (left), with the projection of observation itself $\bmx_i$ (for $i=1,...,n$) in bold on each optimal direction.

This first plot confirms correctness of calculation of the (projection) depth as we see the projections of points $\bmx_i$ (red points) are more outside for lower depth values. Further, we observe apartness of the $10$ anomalies: depth-sorted indices of their optimal directions are $1$--$10$ on ordinate since they possess lowest depth.

Second plot (Figure~\ref{fig:explanation:prj}, right) depicts heat-map of scalar products between all pairs of optimal directions, with their indices again sorted increasing due to depth values. One observes very high values for optimal directions of the ten anomalies, which indicates that all these ten optimal directions are very close to each other: thus the ten anomalies lie in the same direction from normal data, and with high probability in a cluster (which is the case, since they have close depth values).

It is important to note that Figure~\ref{fig:explanation:prj} can serve as a visualization-explanation tool for space of any dimension $\mathbb{R}^d$.

\begin{figure}[!t]
    \centering
        \includegraphics[trim=0cm 0.25cm 0cm 0.5cm,clip=true,scale=.75]{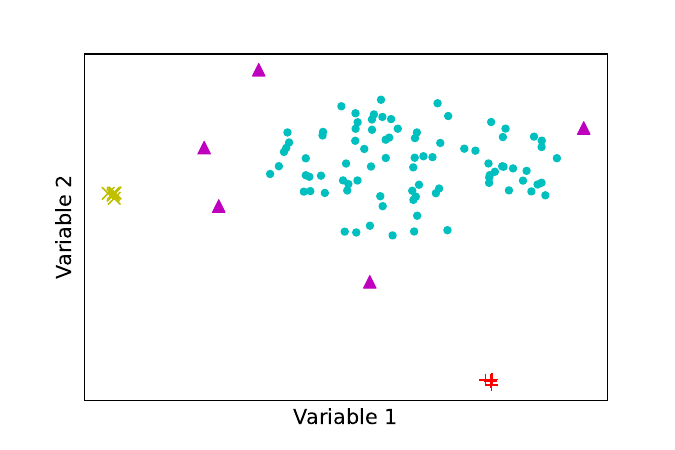}
    \caption{$2-$dimensional visualization of the setting used in Figure~\ref{fig:explanation:prj20}.}\label{fig:explanation:prj20_set}
\end{figure}

\begin{figure}[!t]
    \begin{center}
        \includegraphics[trim=0cm 0cm 1cm 0.5cm, clip=true, scale=.55]{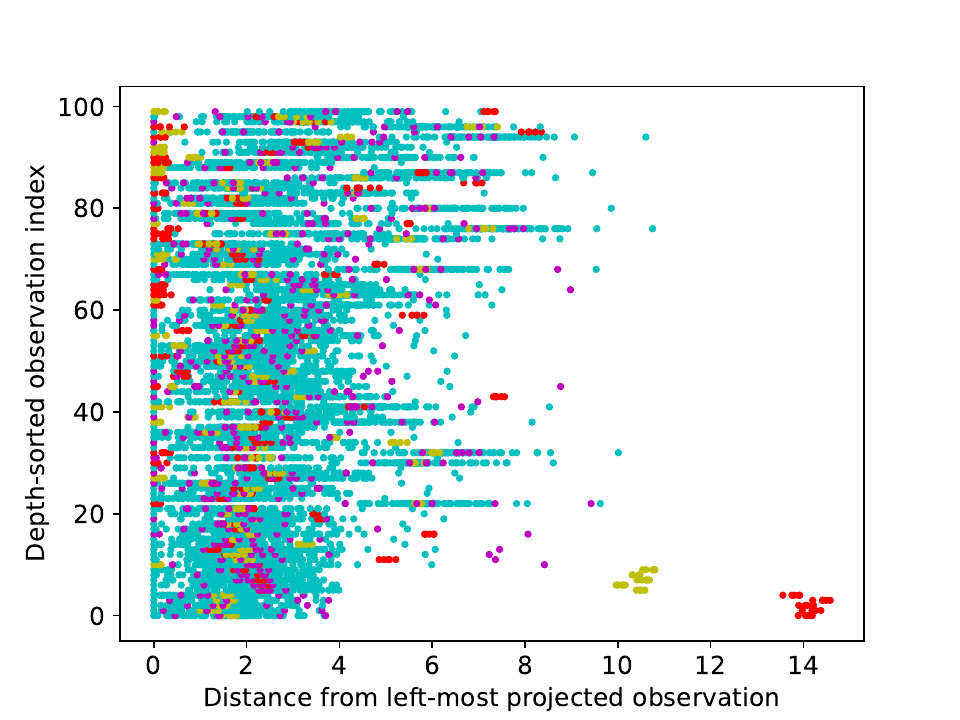}\,\includegraphics[trim=1cm 0cm 1cm 0.5cm, clip=true, scale=.55]{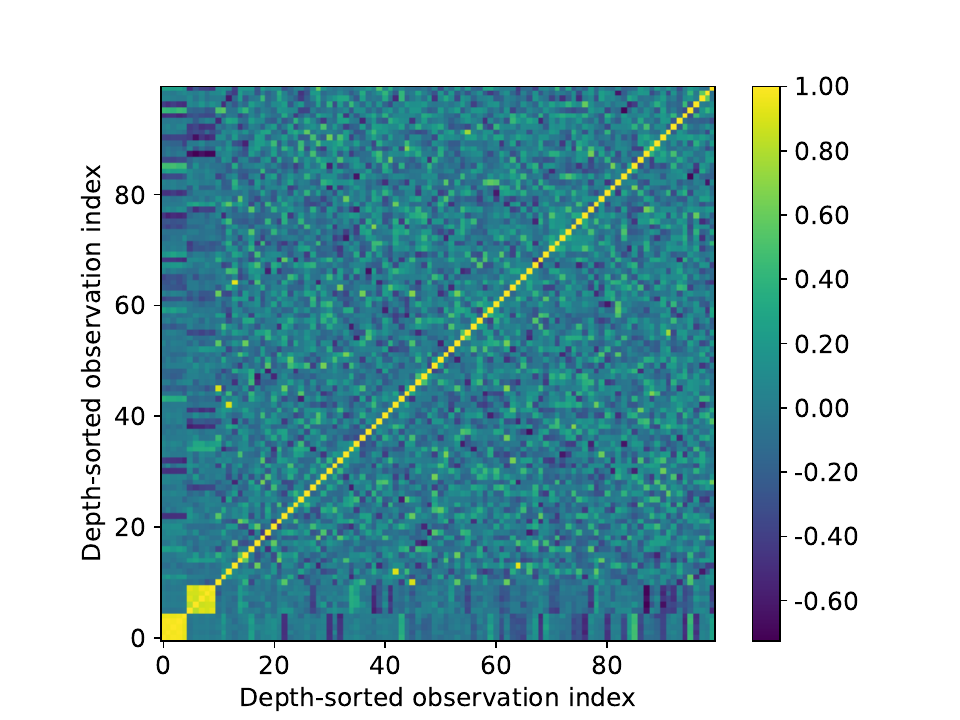}
    \end{center}
    \caption{Example with $d=20$, 2 clusters and isolated anomalies.}\label{fig:explanation:prj20}
\end{figure}

Another setting is explored in Figure~\ref{fig:explanation:prj20} in higher dimension and with a contamination represented (only for visualization) in Figure~\ref{fig:explanation:prj20_set}. The same setting found in Section~\ref{sec:suitability} is used but we increased dimensions to $20$ and added a second anomaly cluster (yellow ``$\times$'') to the first (red ``$+$''). They are placed in the two last principal components obtained from normal data. Finally we placed some isolated anomalies (magenta ``$\blacktriangle$'') sampled from elliptical Cauchy distribution and placed at a distance between normal points and both anomaly clusters.

It is shown on left Figure~\ref{fig:explanation:prj20}, the distinction between the three sources of abnormal points (red and yellow clusters and magenta isolated ones) on different level of abnormality (measured by depth in ascending order). But now, the additional plot on right Figure~\ref{fig:explanation:prj20} shows only the two clusters with optimal directions near each other. This actually gives us additional information about nature of these anomalies. 

\section{Computational tractability}\label{sec:comptract}

In this section, we shall explore computational properties of data depths from two points of view: numerical and statistical. As we have seen in Section~\ref{ssec:depths:comp}, data depth notions possessing most attractive (for anomaly detection purposes) properties (\textit{e.g.}, affine invariance, robustness) demand computational time exponentially increasing with space dimension for exact calculation. This impedes their application in dimensions of order $50$ or $10$, and, depending on data set size, even $5$. The proposed solution is using approximate computation to reasonably sacrifice precision for dramatic decrease in computational time. The goal here is to develop more intuition about the trade-off between computational time and precision of anomaly detection. As we shall see, even if data depth is computed only approximately, this can be sufficient to identify maximum (and up to all, depending on setting) of abnormal observations.

The following distributional setting shall be used in the rest of this section: data set $\bmX_{tr}$ consisting of $n=1000$ observations and containing $5\%$ of anomalies. Normal data are generated from multivariate Gaussian distribution centered in the origin with Toeplitz covariance matrix. Anomalies are drawn from multivariate Gaussian distribution with covariance matrix $\boldsymbol{I}_{d \times d} / 10$ and located $1.25$ Mahalanobis distances from the origin in direction of the smallest principal vector of normal data. This is a relatively difficult setting requiring depth-computation precision and not allowing to identify all anomalies for higher dimensions.

\begin{figure}[!t]
	\begin{center}
	\begin{tabular}{cc}
		$d = 10$ & $d = 50$ \\
		\includegraphics[trim=0.5cm 0cm 1cm 1cm,clip=true,scale=.53]{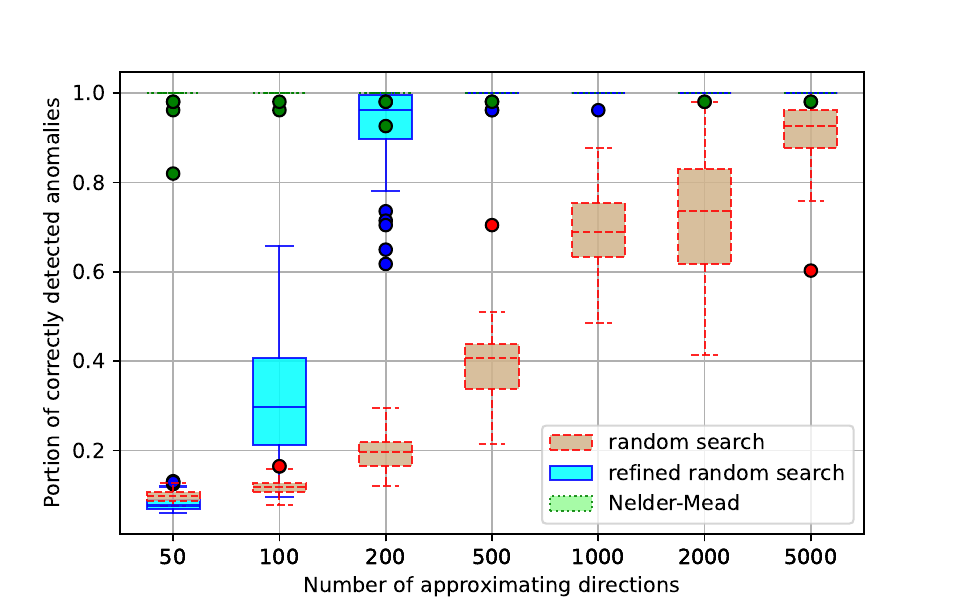} & \includegraphics[trim=0.25cm 0cm 1cm 1cm,clip=true,scale=.53]{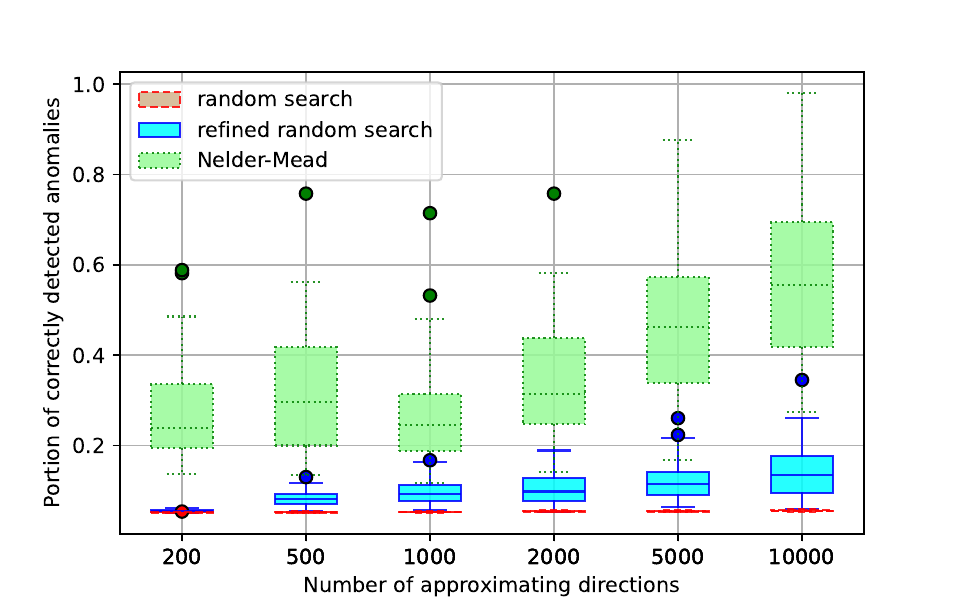}
	\end{tabular}
	\end{center}
	\caption{Boxplots for measure~\eqref{equ:critboxplot} over $50$ repetitions for the distribution from Section~\ref{sec:comptract} (for $n=1000$ observations containing $5\%$ of anomalies) for random search, refined random search and Nelder-Mead approximation algorithms from~\cite{DyckerhoffMN21}. Left: in $\mathbb{R}^{10}$. Right: in $\mathbb{R}^{50}$.}\label{fig:compdirs}
\end{figure}

First, based on~\cite{DyckerhoffMN21}, let us explore the numerical aspect of approximation, \textit{i.e.}, how precision of anomaly detection depends on time (represented by the number of used directions). For $d=10,50$, we measure~\eqref{equ:critboxplot} over $50$ repetitions and visualize it in boxplots of Figure~\ref{fig:compdirs}, approximating projection depth~\eqref{equ:depthPrj} using \textit{random search} (RS), \textit{refined random search} (RRS) and \textit{Nelder-Mead} (NM) algorithms, with the last two containing elements of optimization. Thus, this experiment shall also illustrate advantage of optimization-involving approximation over purely random one, with the last one suffering from curse of dimension as it has been shown by~\cite{NagyDM20}. As we can see from Figure~\ref{fig:compdirs} (left), depth-based anomaly detection rule copes with the task for $d=10$ perfectly even with small number of directions when using RRS approximation (\textit{e.g.}, with $200$ directions depth calculation for all $1000$ points of one data set took less than $20$ seconds on a single core of Apple M1 Max chip). RS algorithm improves with growing number of directions (for $5000$ directions the same calculation took up to $500$ seconds).

When increasing dimension to $d=50$, the task of anomaly detection becomes difficult to tackle since due to growing distances the cluster of abnormal observations becomes unrecognizable. On the other hand this allows to benchmark the two algorithms along the entire abscissa, see Figure~\ref{fig:compdirs} (right). It is noteworthy that such situations are not rare in practice and still beneficial: \textit{e.g.}, when searching for inconsistencies in enterprise's transactions (the operation to be done by hand on up to millions records) narrowing down search of few anomalies to hundreds transactions facilitates dramatically the audit work. The difference between approximation methods becomes even more visible, while purely random RS does not cope with the task, optimization NM starts to get interesting results considering the size of $d$.

\begin{figure}[!t]
	\begin{center}
	\begin{tabular}{cc}
		$d = 10$ & $d = 20$ \\
		\includegraphics[trim=0.5cm 0cm 1cm 1cm,clip=true,scale=.57]{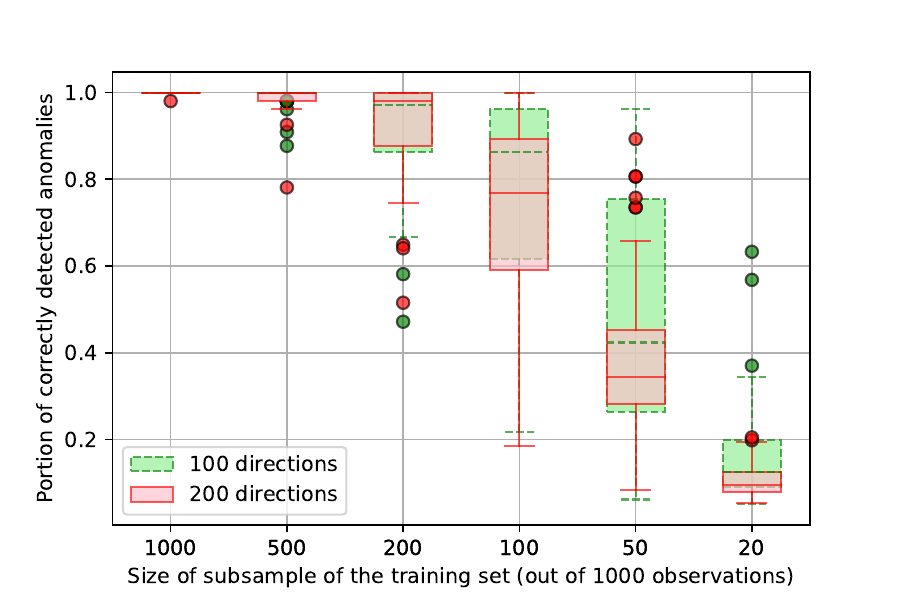} & \includegraphics[trim=0.25cm 0cm 1cm 1cm,clip=true,scale=.57]{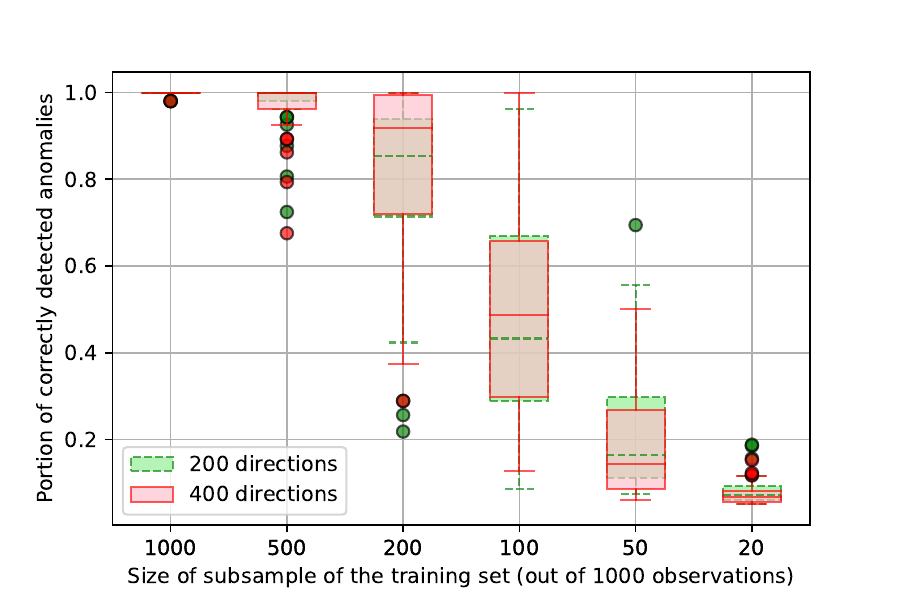}
	\end{tabular}
	\end{center}
	\caption{Boxplots for measure~\eqref{equ:critboxplot} over $50$ repetitions for the distribution from Section~\ref{sec:comptract} (for $n=1000$ observations containing $5\%$ of anomalies) for Nelder-Mead approximation algorithm from~\cite{DyckerhoffMN21}. For two numbers of approximating directions (linearly increasing with dimension), the sub-sample of $\bmX_{tr}$ with respect to which the depth of each observation $\in\bmX_{tr}$ is computed is gradually reduced. Left: in $\mathbb{R}^{10}$. Right: in $\mathbb{R}^{20}$.}\label{fig:compsubsample}
\end{figure}

Second, we study the statistical aspect and try to increase computation speed by sub-sampling. Here, we computed depth of each point of $\bmX_{tr}$ with respect to its random subset, and repeat the experiment $50$ times, employing the Nelder-Mead algorithm as described in~\cite{DyckerhoffMN21}. For two different numbers of directions in each case, for $d=10,20$, \eqref{equ:critboxplot} is indicated in Figure~\ref{fig:compsubsample}. Clearly, when reducing the size of the data set used the quality of anomaly detection decreases, while this does not happen immediately suggesting a possible compromise. Indeed, in the problems where the data set is too large, sub-sampling is not expected to substantially reduce the quality, whereas for a small data set computation should be fast enough even without sub-sampling.

While continuing the example with projection depth, as experimental evidence of~\cite{DyckerhoffMN21} suggests, the results from the first experiment here are (accounting for robustness and anomalies' configuration) extendable to the multitude of depths satisfying the projection property~\citep{Dyckerhoff04}. The conclusions of the second experiment are even more general.



\section{Real data sets}

\begin{table}[htbp]
\begin{center}
\scalebox{0.70}{
\begin{tabular}{llrrrrrrrrr}
\toprule
Name & Size & Dimension & \% Anomaly & AE L1 & AE MSE & Deep SVDD & IF & LOF & OCSVM & PD \\
\midrule
ALOI & - & 27 & 3.04 & 10.1 & 11.0 & 9.2 & 5.5 & \textbf{21.1} & 9.2 & 6.4 \\
annthyroid & 7200 & 6 & 7.42 & 43.8 & 39.8 & 62.5 & 60.2 & 58.0 & 23.3 & \textbf{77.8} \\
breastw & 683 & 9 & 34.99 & \textbf{100.0} & \textbf{100.0} & \textbf{100.0} & \textbf{100.0} & \textbf{100.0} & \textbf{100.0} & \textbf{100.0} \\
cardio & 1831 & 21 & 9.61 & \textbf{100.0} & \textbf{100.0} & 24.1 & \textbf{100.0} & 43.1 & 65.5 & 93.1 \\
Cardiotocography & 2114 & 21 & 22.04 & 92.9 & \textbf{93.5} & 69.5 & 86.4 & 78.6 & 73.4 & 86.4 \\
celeba & - & 39 & 2.24 & \textbf{33.8} & \textbf{33.8} & 10.0 & 23.8 & 3.8 & 5.0 & 23.8 \\
cover & - & 10 & 0.96 & 28.6 & 35.7 & 14.3 & 25.0 & \textbf{78.6} & 0.0 & 10.7 \\
donors & - & 10 & 5.93 & \textbf{54.7} & 15.3 & 22.1 & 37.9 & 5.8 & 4.2 & 41.1 \\
fault & 1941 & 27 & 34.67 & \textbf{100.0} & \textbf{100.0} & \textbf{100.0} & \textbf{100.0} & \textbf{100.0} & \textbf{100.0} & \textbf{100.0} \\
fraud & - & 29 & 0.17 & 66.7 & 66.7 & \textbf{100.0} & 66.7 & \textbf{100.0} & 50.0 & 66.7 \\
glass & 214 & 7 & 4.21 & \textbf{33.3} & \textbf{33.3} & 0.0 & \textbf{33.3} & \textbf{33.3} & 0.0 & \textbf{33.3} \\
Hepatitis & 80 & 19 & 16.25 & \textbf{75.0} & \textbf{75.0} & 50.0 & 50.0 & 50.0 & 50.0 & 25.0 \\
http & - & 3 & 0.39 & \textbf{100.0} & \textbf{100.0} & \textbf{100.0} & \textbf{100.0} & \textbf{100.0} & \textbf{100.0} & \textbf{100.0} \\
Ionosphere & 351 & 32 & 35.90 & \textbf{100.0} & \textbf{100.0} & \textbf{100.0} & \textbf{100.0} & \textbf{100.0} & \textbf{100.0} & \textbf{100.0} \\
landsat & 6435 & 36 & 20.71 & 44.3 & 44.5 & \textbf{77.0} & 57.5 & 64.1 & 55.7 & 65.0 \\
letter & 1600 & 32 & 6.25 & 24.2 & 33.3 & 39.4 & 27.3 & 57.6 & \textbf{78.8} & 33.3 \\
Lymphography & 148 & 18 & 4.05 & \textbf{100.0} & \textbf{100.0} & \textbf{100.0} & \textbf{100.0} & \textbf{100.0} & \textbf{100.0} & 50.0 \\
magic.gamma & - & 10 & 35.16 & \textbf{100.0} & \textbf{100.0} & \textbf{100.0} & \textbf{100.0} & \textbf{100.0} & \textbf{100.0} & \textbf{100.0} \\
mammography & - & 6 & 2.32 & \textbf{55.0} & 51.2 & 43.8 & 50.0 & 47.5 & 31.2 & 38.8 \\
optdigits & 5216 & 64 & 2.88 & 0.0 & 0.0 & 0.0 & \textbf{12.0} & 6.0 & 8.0 & 2.0 \\
PageBlocks & 5393 & 10 & 9.46 & 86.3 & \textbf{88.7} & 36.3 & 85.1 & 73.2 & 28.0 & 75.0 \\
pendigits & 6870 & 16 & 2.27 & \textbf{57.7} & 34.6 & 11.5 & 53.8 & 7.7 & 32.7 & 15.4 \\
Pima & 768 & 8 & 34.90 & \textbf{100.0} & \textbf{100.0} & \textbf{100.0} & \textbf{100.0} & \textbf{100.0} & \textbf{100.0} & \textbf{100.0} \\
satellite & 6435 & 36 & 31.64 & 95.1 & 93.9 & 97.9 & 97.5 & 93.2 & 98.1 & \textbf{98.2} \\
satimage-2 & 5803 & 36 & 1.22 & 91.3 & 95.7 & 4.3 & 95.7 & 8.7 & 0.0 & \textbf{100.0} \\
shuttle & - & 9 & 7.15 & 98.0 & 98.4 & 35.1 & 99.2 & 44.4 & 45.6 & \textbf{100.0} \\
skin & - & 3 & 20.75 & 95.4 & 89.0 & 76.6 & \textbf{100.0} & 60.3 & 41.7 & \textbf{100.0} \\
smtp & - & 3 & 0.03 & \textbf{0.0} & \textbf{0.0} & \textbf{0.0} & \textbf{0.0} & \textbf{0.0} & \textbf{0.0} & \textbf{0.0} \\
SpamBase & 4207 & 57 & 39.91 & \textbf{100.0} & \textbf{100.0} & \textbf{100.0} & \textbf{100.0} & \textbf{100.0} & \textbf{100.0} & \textbf{100.0} \\
Stamps & 340 & 9 & 9.12 & \textbf{100.0} & 90.0 & 50.0 & \textbf{100.0} & 50.0 & 40.0 & 80.0 \\
thyroid & 3772 & 6 & 2.47 & 80.6 & 67.7 & 0.0 & \textbf{93.5} & 77.4 & 51.6 & \textbf{93.5} \\
vertebral & 240 & 6 & 12.50 & 20.0 & 20.0 & 20.0 & 30.0 & \textbf{40.0} & 30.0 & 10.0 \\
vowels & 1456 & 12 & 3.43 & 29.4 & 29.4 & 23.5 & 47.1 & \textbf{82.4} & 47.1 & 52.9 \\
Waveform & 3443 & 21 & 2.90 & 12.1 & 12.1 & 15.2 & 18.2 & \textbf{48.5} & 39.4 & 12.1 \\
WBC & 223 & 9 & 4.48 & \textbf{100.0} & \textbf{100.0} & 33.3 & \textbf{100.0} & \textbf{100.0} & \textbf{100.0} & \textbf{100.0} \\
WDBC & 367 & 30 & 2.72 & \textbf{100.0} & \textbf{100.0} & 33.3 & \textbf{100.0} & \textbf{100.0} & 0.0 & \textbf{100.0} \\
Wilt & 4819 & 5 & 5.33 & 1.2 & 2.4 & 12.9 & 2.4 & 38.8 & 21.2 & \textbf{54.1} \\
wine & 129 & 13 & 7.75 & \textbf{100.0} & \textbf{100.0} & 33.3 & \textbf{100.0} & \textbf{100.0} & \textbf{100.0} & \textbf{100.0} \\
WPBC & 198 & 33 & 23.74 & 75.0 & 75.0 & 75.0 & 81.2 & 81.2 & 68.8 & \textbf{87.5} \\
yeast & 1484 & 8 & 34.16 & \textbf{100.0} & \textbf{100.0} & \textbf{100.0} & \textbf{100.0} & \textbf{100.0} & \textbf{100.0} & \textbf{100.0} \\
\midrule
Top Score & . & . & . & 20 & 18 & 11 & 18 & 19 & 12 & 20 \\
Top Unique & . & . & . & 3 & 2 & 1 & 1 & 5 & 1 & 6 \\
\bottomrule
\end{tabular}
}
\end{center}
\caption{Performances on real data sets for AE (MSE and L1 losses), Deep SVDD, IF, LOF, OCSVM and PD. The metric used is precision (also called positive predictive value) between 0 and 100\%. When size is -, only 10~000 observations were sampled.}\label{tab:datasets}
\end{table}

To shed more light on performance of the depth-based anomaly detection approach in a general context, we consider $40$ real-world data sets \citep[][known as the ODDS library]{Rayana2016} in a comparative study. We thus contrast data depth with autoencoder (using both $L_1$ and $L_2$ losses), deep SVDD, IF, OC-SVM, and LOF. For each data set, we use a stratified split (to preserve contamination rate) with proportion $66\%$/$33\%$ for train and test data respectively. We assess the methods' performance using the true contamination rate $\bmc \in [0,1]$ to compute the precision (or positive predictive rate) and indicate the portion of observations correctly labeled as abnormal among the proportion $\bmc$ of highest anomaly score (or lowest depending on the method). The results are indicated in Table~\ref{tab:datasets}.

With the goal of this article being not to illustrate superior performance of depth-based anomaly detection, but to indicate its place in machine learning, the results are satisfactory. Data depth score best in $21$ out of $40$ data sets, ranging in dimension from $3$ to $64$ and having up to $10\,000$ observations. Furthermore, for such data sets as ``annthyroid'', ``Wilt'' or ``WPBC'' it substantially outperforms the state of the art.

\section{Outlook}\label{sec:outlook}

Data depth has undergone substantial theoretical developments during a few recent decades and possesses attractive properties, such as non-parametricity, robustness, affine invariance, \textit{etc}. After having been computationally reachless for long time, more recently efficient exact and approximate computation methods have been proposed. Together with previous theoretical results, these transform data depth into a powerful tool for anomaly detection.

When applying data depth for anomaly detection, several aspects should be taken into account: these were addressed in sections of the present article. Section~\ref{sec:algorithmic} analyses computational aspects of data depth. In Section~\ref{sec:suitability}, on an example of several depth notions, importance of robustness, affine invariance, non-vanishing beyond data's convex hull is underlined. In certain (not uncommon) settings highlighted in Section~\ref{sec:advantages}, data depth competitively compares with such widely used anomaly detection tools as autoencoder, local outlier factor, one-class support vector machine, and isolation forest, additionally providing explainability. Experiments of Section~\ref{sec:comptract} illustrate that---with reasonably limited resources---anomaly detection can be performed in relatively high dimensions when properly choosing the degree of approximation and computed data depth with respect to a sub-sample only.

Even though the topic itself deserves a separate work, it is noteworthy that experimental evidence from the current article can be useful for anomaly detection in functional data. When employing (a selected notion of) functional data depth, anomaly detection rule~\eqref{equ:projdepth} can be readily applied to functions (or time series), given the threshold is properly chosen. Furthermore, in case if a notion of integrated functional depth is used, for each argument value, multivariate depth is to be considered. Already from computational perspective, the ideas of Section~\ref{sec:comptract} can be applied as well.

It is important to conclude with the following \textit{disclaimer}: The presented in the current article examples were designed to illustrate advantages of depth-based anomaly detection and their immediate generalization can be limited.

The source codes (in \texttt{Python}) of all examples and experiments contained in the current article can be downloaded from the author's website.

\section*{Acknowledgements}

The authors are grateful for the insightful remarks of Prof. Florence d'Alch\'{e}-Buc on the manuscript. The authors further  gratefully acknowledge the support of the Young Researcher Grant of the French National Agency for Research (ANR JCJC 2021) in category Artificial Intelligence registered under the number ANR-21-CE23-0029-01 and the support of the CIFRE grant number 2021/1739.


\section*{Appendix: experiment~\ref{ssec:adv:extrap} with higher dimension}\label{app:extrap:10:20}

For both training and testing sets with $d=10$ and $d=20$, anomaly score (on ordinate) by LOF (top left), OCSVM (top right), IF (bottom left) and data depth (bottom right). Separated by vertical lines from left to right are  normal observations (indices on abscissa $1$-$225$, blue dots) and anomalies ($226$-$250$, red pluses) of the training set, normal observations ($251$-$750$, bigger cyan dots) and anomalies ($751$-$1000$, magenta crosses) of the test set.

\begin{figure}[ht]
	\centering
        \includegraphics[trim=0cm 0cm 1cm 1cm, clip=true, scale=.475]{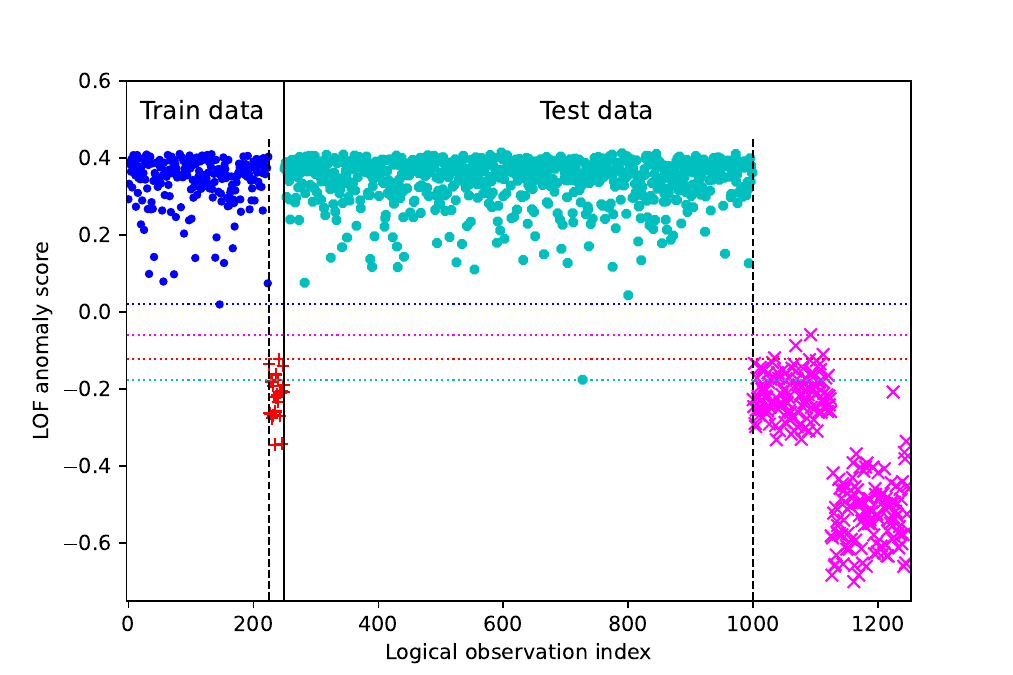}\,\includegraphics[trim=0cm 0cm 1cm 1cm, clip=true, scale=.475]{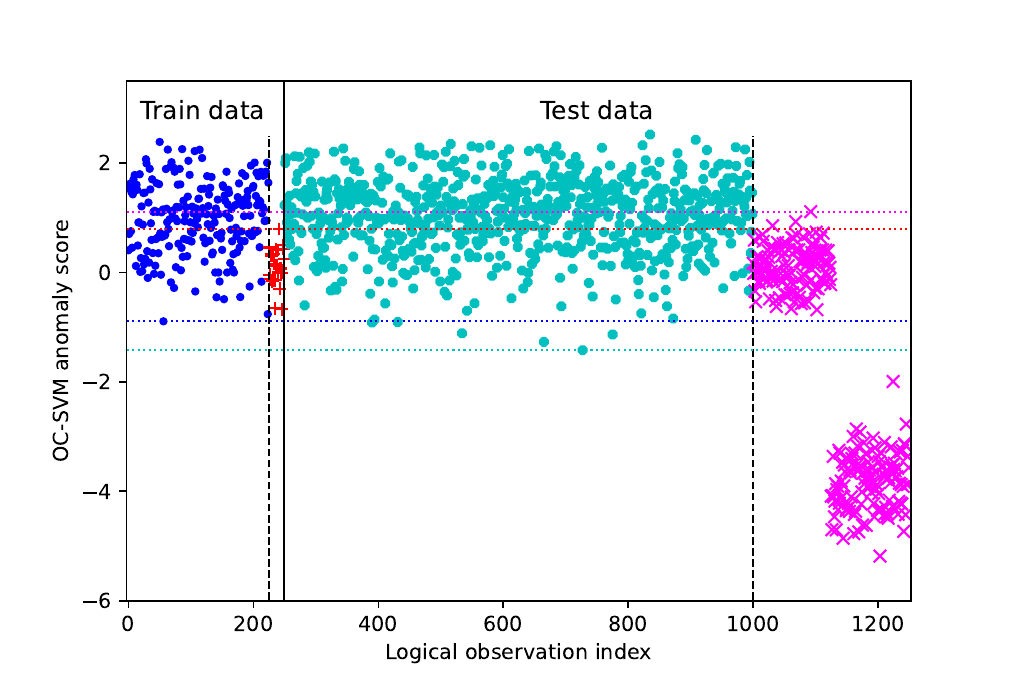} \\
        \includegraphics[trim=0cm 0cm 1cm 1cm, clip=true, scale=.475]{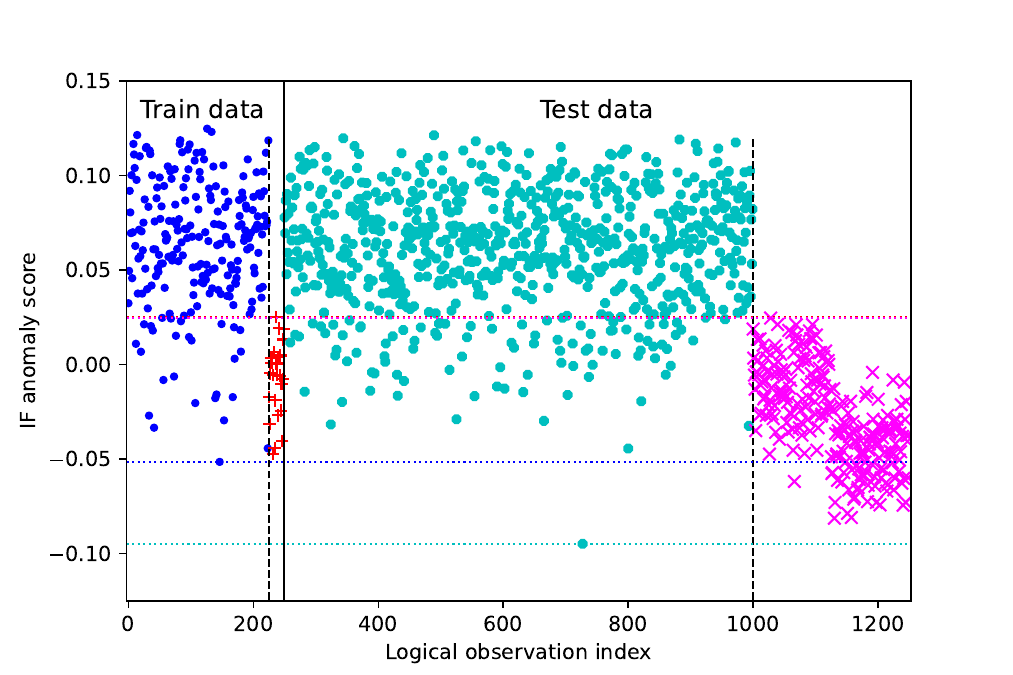}\,\includegraphics[trim=0cm 0cm 1cm 1cm, clip=true, scale=.475]{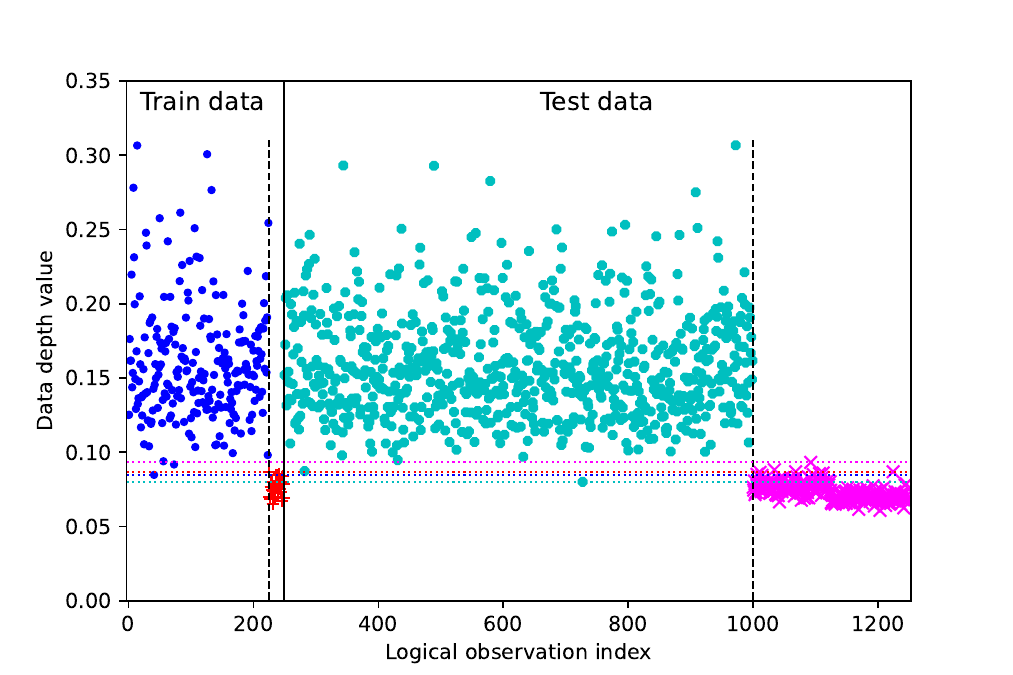}
    \caption{Experiment with $d=10$}
\end{figure}

\begin{figure}[ht]
    \centering
        \includegraphics[trim=0cm 0cm 1cm 1cm, clip=true, scale=.475]{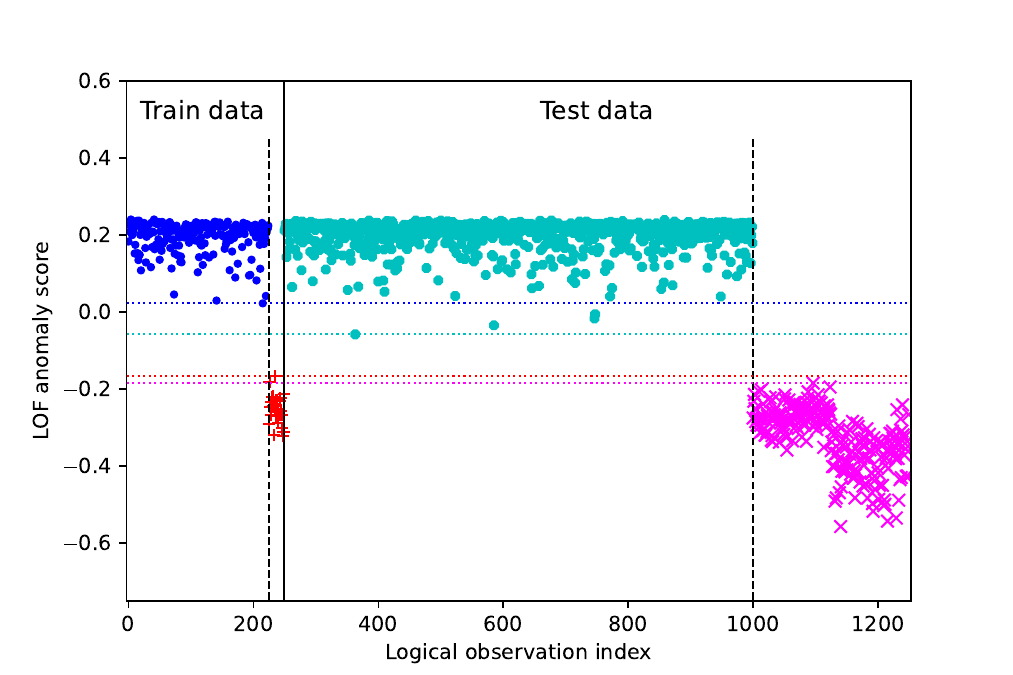}\,\includegraphics[trim=0cm 0cm 1cm 1cm, clip=true, scale=.475]{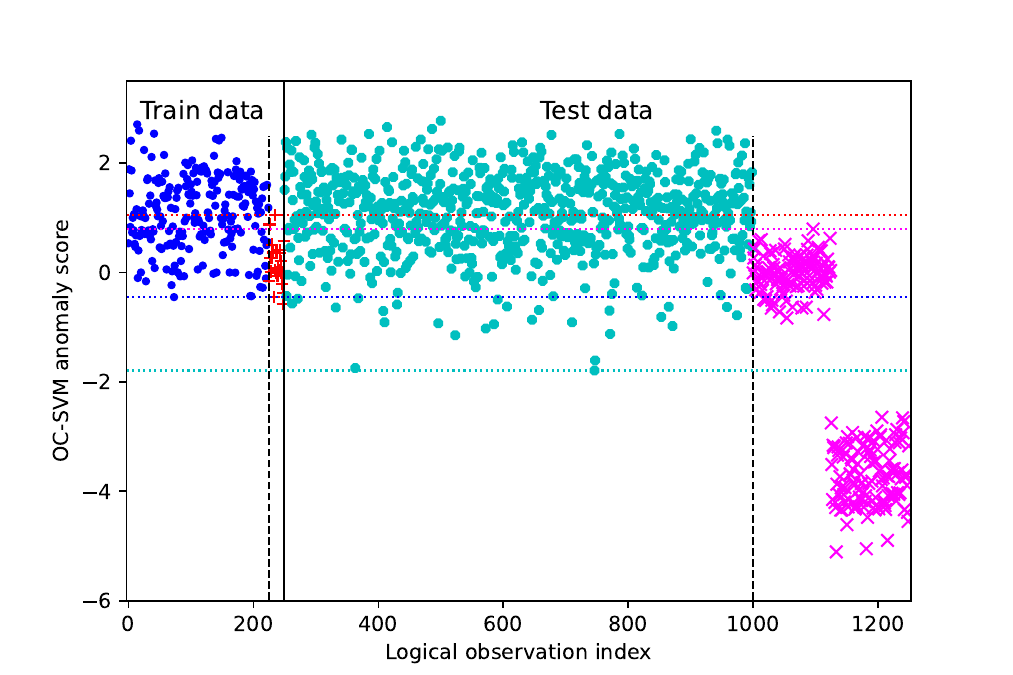} \\
        \includegraphics[trim=0cm 0cm 1cm 1cm, clip=true, scale=.475]{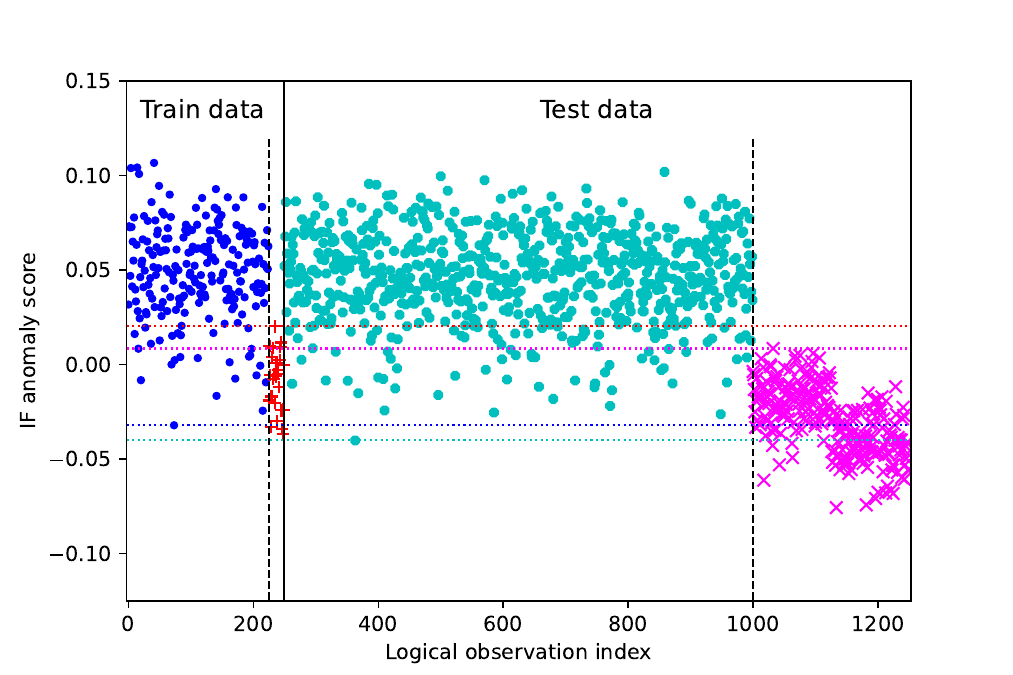}\,\includegraphics[trim=0cm 0cm 1cm 1cm, clip=true, scale=.475]{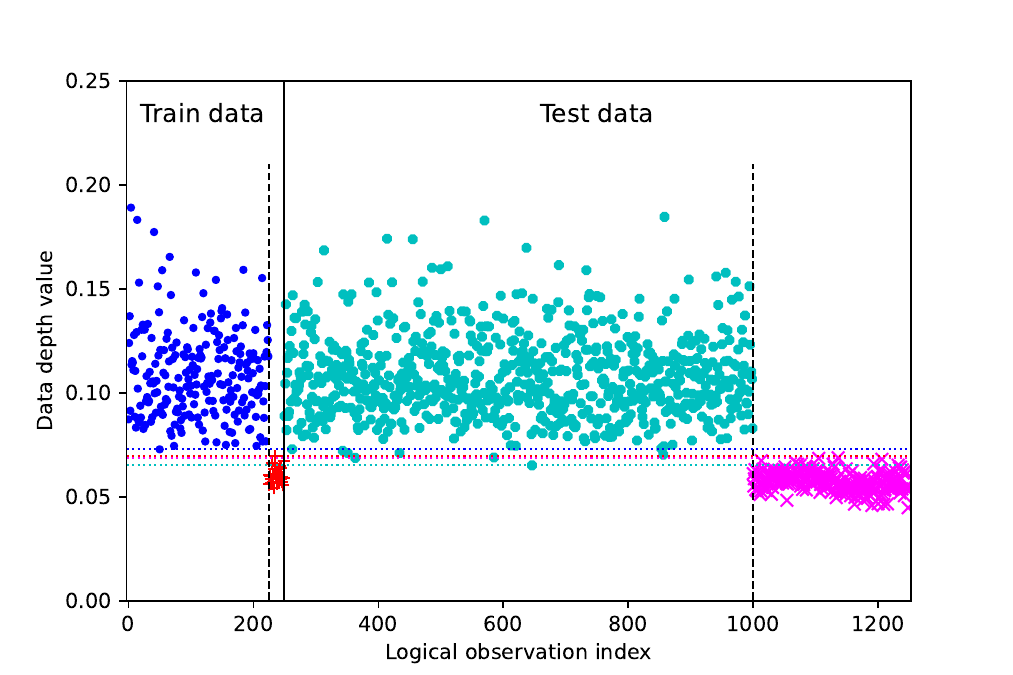}
    \caption{Experiment with $d=20$}
\end{figure}

\end{document}